\documentclass[10pt,journal,compsoc]{IEEEtran}
\pdfoutput=1 
\usepackage[backref]{hyperref} 
\usepackage{bm}
\usepackage{array}
\usepackage{amssymb}
\usepackage{amsmath}
\usepackage{algorithmicx}
\usepackage[linesnumbered, ruled]{algorithm2e}
\usepackage{graphicx}
\usepackage{subfigure}
\usepackage{multirow}
\usepackage{makecell}
\ifCLASSOPTIONcompsoc
  \usepackage[nocompress]{cite}
\else
  \usepackage{cite}
\fi

\ifCLASSINFOpdf
\else
\fi

\begin{document}

\title{A New Outlier Removal Strategy Based on Reliability of Correspondence Graph for Fast Point Cloud Registration}

\author{Li Yan, Pengcheng Wei, Hong Xie, Jicheng Dai, Hao Wu, Ming Huang
\IEEEcompsocitemizethanks{\IEEEcompsocthanksitem L. Yan, P. Wei, H. Xie, J. Dai and H. Wu are with the School of Geodesy and Geomatics, Wuhan University, Wuhan 430079, China.\protect\\

E-mail: lyan@sgg.whu.edu.cn; wei.pc@whu.edu.cn;hxie@sgg.whu.edu.cn;
dai-jicheng@whu.edu.cn; 2021202140055@whu.edu.cn 
\IEEEcompsocthanksitem M. Huang. is with the School of Geomatics and Urban Spatial Infor-mation, Beijing University of Civil Engineering and Architecture, Bei-jing 100044, China.\protect\\
E-mail:  huangming@bucea.edu.cn
\IEEEcompsocthanksitem Corresponding author: Pengcheng Wei and Hong Xie.

}
\thanks{}}

%

\IEEEtitleabstractindextext{%
\begin{abstract}
  Registration is a basic yet crucial task in point cloud processing. In correspondence-based point cloud registration, matching correspondences by point feature techniques may lead to an extremely high outlier ratio. Current methods still suffer from low efficiency, accuracy, and recall rate. We use a simple and intuitive method to describe the 6-DOF (degree of freedom) curtailment process in point cloud registration and propose an outlier removal strategy based on the reliability of the correspondence graph. The method constructs the corresponding graph according to the given correspondences and designs the concept of the reliability degree of the graph node for optimal candidate selection and the reliability degree of the graph edge to obtain the global maximum consensus set. The presented method could achieve fast and accurate outliers removal along with gradual aligning parameters estimation. Extensive experiments on simulations and challenging real-world datasets demonstrate that the proposed method can still perform effective point cloud registration even the correspondence outlier ratio is over 99\%, and the efficiency is better than the state-of-the-art. Code is available at https://github.com/WPC-WHU/GROR.
\end{abstract}

}

\maketitle

\IEEEdisplaynontitleabstractindextext

%
\IEEEpeerreviewmaketitle

\IEEEraisesectionheading{\section{Introduction}\label{sec:introduction}}

%
%
%
%
\IEEEPARstart{T}{he} point cloud records the 3D information and is one of the most suitable data to show the objective world. With the rapid development of 3D laser scanning and computer technology, point cloud has been widely applied in many fields, such as computer vision \cite{[1]}, photogrammetry \cite{[2]}, forest survey \cite{[3]}, smart city \cite{[4]}, robotics \cite{[5]}, etc. However, the point cloud scanned by one station is usually incomplete and can not obtain all the required information. To improve the integrity and quality of point cloud data, it is necessary to integrate multi-angle, multi-platform, multi-station, and multi-temporal scanning data into the same coordinate system through registration technology \cite{[6]}. Suppose there are two point clouds $\mathcal{P}$(target) and   $\mathcal{Q}$(source), their coordinate systems are ${O_{\cal P}} - {X_{\cal P}}{Y_{\cal P}}{Z_{\cal P}}$ and ${O_{\cal Q}} - {X_{\cal Q}}{Y_{\cal Q}}{Z_{\cal Q}}$ respectively. The classical coordinate system transformation is completed by the Bursa-Wolf model \cite{[7]} according to the Euler angle. The model adopts seven parameters, including three rotation parameters ${\varepsilon _X},{\varepsilon _Y},{\varepsilon _Z}$, three translation parameters $\Delta X,\Delta Y,\Delta Z$, and one scale parameter $\delta u$. To solve the seven parameters, three non-collinear correspondence points are needed at least. When there are more than three correspondences, the most probable value of the transformation parameters  ${\bf{R}},{\bf{t}}$ can be estimated by the least-square method \cite{[8]}. Obviously, obtaining accurate correspondences is very crucial for registration. ICP \cite{[9]} algorithm takes the nearest points as the correspondence and iteratively optimizes the transformation parameters until the accuracy meets requirements. However, the algorithm needs initial parameters, otherwise, it is easy to fall into the local optimum \cite{[10]}.

In order to identify a point between laser scans more clearly, many descriptors based on the geometric characteristic histogram are designed, such as FPFH \cite{[11]}, SHOT \cite{[12]}, BSC \cite{[13]}, etc. The descriptor-based correspondence matching method can effectively match the point pair with similar features. However, there are also a large fraction of outliers in the initial correspondences. Two main difficulties appear in the research of outlier removal: 1) The features of different point cloud scenes are diverse, the applicability of current methods is limited due to the merely simple geometric constraints (such as distance and angle) are used. Usually, one method often can not complete the task of outlier removal of multiple type scenarios which leads to low robustness and adaptability; 2) When there is a great number of correspondences to be checked or extremely high outlier ratios, the effect of outlier removal is difficult to be satisfactory, which affects the accuracy of registration. In addition, the classic removal strategies usually need a lot of trails, resulting in low efficiency. In order to solve the above problems, a robust outlier removal strategy based on the reliability of the correspondence graph is proposed for fast point cloud registration. The main contribution of this paper includes:

\begin{itemize}
  \item A simple and intuitive method is designed to describe the curtailment process of the degree of freedom in point cloud registration. According to this process, an outlier removal strategy based on the reliability correspondence graph is proposed and we named it GROR (graph reliability outlier removal).
  \item Two undirected complete graphs are constructed according to the correspondences. Then, we propose the concept of reliability degree of graph nodes which can be measured by the adjacency matrix ${\cal A}$ of the graph. According to the reliability degree, we select $K$ correspondences with high reliability as the optimal candidates for further processing. 
  \item An edge-node affinity matrix ${\cal M}$ is defined to measure the reliability degree of graph edges in the same constraint function space. A loose constraint function ${{\cal F}_1}$ and a compact constraint function ${{\cal F}_2}$ are designed to compare the reliability of the correspondence edge to accelerate the algorithm and obtain the maximum consensus set of the corresponding point.
\end{itemize}

\section{RELATED WORK}\label{sec:related work}
As a basic but challenging task in the fields of photogrammetry, computer vision, and robotics, point cloud registration has been widely studied and applied. The current point cloud registration technology generally adopts the registration strategy from coarse to fine. The mode of fine registration has basically been fixed to use the ICP algorithm \cite{[9]} and its variants \cite{[14],[15]}. Since the ICP needs a proper initial transformation matrix, when the relative position between the two point clouds is completely unclear, calculating the approximate transformation matrix through the coarse registration becomes crucial for the registration technology. The proposed GROR belongs to the coarse registration technology based on correspondence. In correspondence-based registration research, two steps are vital, one is correspondence matching, and the other is outlier removal.

\subsection{Correspondence matching}\label{sec:correspondece matching}
The random sampling consensus (RANSAC) \cite{[14]} algorithm is one of the earliest methods applied to correspondence matching. It randomly selects three pairs of points from ${\cal P}$ and ${\cal Q}$ to calculate the transformation matrix and the registration score, then it repeats the sampling and test process until the largest registration score is found or the highest sampling times is reached \cite{[15]}. RANSAC algorithm requires a large number of iterations to seek a satisfactory solution which results in low efficiency. In order to reduce the attempts to establish reliable matches, some methods match correspondences according to geometric constraints \cite{[16]}. The most representative is 4PCS \cite{[17]}. Based on the RANSAC framework, the 4PCS constructs the non-collinear fourpoint congruent sets in point clouds ${\cal P}$ and ${\cal Q}$, and uses affine invariance to find similar correspondence pairs in ${\cal P}$ and ${\cal Q}$. The final transformation matrix is estimated according to optimal four-point congruent which is measured by the overlap of aligned point clouds. However, when the amount of points is huge, enormous four-point congruent are constructed which causes poor efficiency. In order to further reduce the matching primitives, Xu et al. \cite{[18]} took the line feature in the point cloud scene as the matching primitives. Similar to the four-point congruent sets, they completed the 4-DOF urban scene registration by constructing 2-lines congruent sets (2LCS). Xu et al. \cite{[19]} took the plane as the matching primitive and proposed voxel-based 4-plane congruent sets (V4PCS) based on the 4PCS strategy. The improved strategies by line features and planar features can greatly reduce the matching primitives and improve the efficiency and stability of registration. However, the application of these algorithms is limited which generally suitable for urban scenes. Using key points with recognition features instead of the original point cloud for correspondence matching can improve efficiency and reduce the impact of noise and outliers. The key points are generally the points with changed features such as inflection points and intersections \cite{[20]}. The popular key point detection algorithms include DOG \cite{[21]}, Harris \cite{[22]}, ISS \cite{[23]}, LSP \cite{[24]}. Keypoint 4PCS \cite{[25]} is the most commonly used improvement of 4PCS based on key points. The algorithm improves the matching efficiency by extracting DOG key points instead of random sampling points. However, the algorithm needs to estimate the degree of overlap in advance. Inaccurate parameters affect the selection range of 4-point congruent sets and then affect the registration accuracy.

The 3D point feature descriptor can encode the information of geometry relationship between a point with its neighbors, and then it is convenient to calculate the similarity of features between point pairs according to the coding, which can help more clearly find the correspondence. Feature descriptor technology has been widely used in the field of point cloud registration. So far, the most widely used descriptors are fast point feature histogram (FPFH) \cite{[11]}, signature of histogram of orientations (SHOT) \cite{[12]}, rotational project statistics (ROPS) \cite{[26]}, 3D scale-invariant feature transform (3D-SIFT) \cite{[27]}, And rotation-invariant descriptor in the frequency domain (RIDF) \cite{[28]}, etc. With the development of deep learning, learning-based descriptors also show robust feature description ability, such as RPM-Net \cite{[29]}, 3DMatch \cite{[30]}, 3DFeat-Net \cite{[31]}, etc. In addition, line and planar primitives in the scene can also establish descriptors. For example, Wei et al. \cite{[10]} proposed a plane shape descriptor, which can be less affected by noise and occlusion in the point cloud than the point-based descriptor. Finally, correspondences are matched by matching score between each primitive descriptor histogram or directly use KD-Tree to conduct the neighbor query of multi-dimensional features histogram. The methods to calculate the matching score include nearest neighbor distance ratio strategy and chisquare test, etc. \cite{[32]}. Although most of the current descriptors have strong feature description ability, affected by the noise, uneven density, occlusion, and the repeated structure in the point cloud scene the correspondences matched according to the descriptor often contain a large number of mismatches (outliers). Outliers need to be processed correctly to complete the task of point cloud registration.

\subsection{Outlier removal}\label{sec:outlier removal}
At present, there are two mainstream methods to cope with the outliers in the correspondence set. 

One is the not guaranteed removal strategy represented by fast global registration (FGR) \cite{[35]}, it randomly selects 1000 pairs as candidate matches from the correspondence set, a tuple geometric test is used to remove partial outliers, then, they use the Geman McClure as objective function and proposed a global method by combining line process with robust estimation to improve the optimization process. FGR can get a high-precision registration result even if there are still outliers in candidates. Although the efficiency of FGR is very fast, when the outlier ratio is high or the registration scene is complex, the 1000 pairs selected by random may contain less correct correspondence. This will force the optimization function to handle too many outliers, which is easy to cause the wrong result. Similarly, Li et al. \cite{[36]} constructed a topological graph according to correspondences, then proposed an edge voting strategy to remove outliers and proposed a new cost function Cauchy-weighted  ${l_q}$-norm, which is still robust even if the outlier rate exceeds 80\% $ \sim $ 90\%.

Another is the guaranteed outlier removal strategy. This method focuses on how to eliminate all outliers, so it does not need outlier optimization functions that can simplify the whole processing steps. A representative of this method is Gore \cite{[33]}. For each correspondence, Gore first regards it as an inlier and seeks the lower and upper bounds for consensus size. If the upper bound conflicts with the lower bound, the correspondence will be removed as a true outlier \cite{[34]}. Gore can effectively remove all outliers and obtain satisfactory registration results, but the complexity of the algorithm is high which leads to low efficiency. In order to further improve the efficiency, Cai et al. \cite{[35]} took the point cloud obtained by laser scan equipped with compensator as the research target and reduced the 6DOF registration problem to 4DOF, they significantly reduced the candidate correspondence set through a deterministic selection scheme and then a fast branch-and-bound (BnB) algorithm is applied to quickly find the optimal alignment parameters. The algorithm greatly improves the efficiency, but when the number of correspondence is very large, the algorithm is still time-consuming. Yang et al. \cite{[36]} proposed a drastic pruning of outliers method by finding the maximum clique named TEASER++, it used a general graph-theoretic framework to decouple the scale, rotation, and translation estimation. The scale and translation estimation are solved via an adaptive voting scheme and the rotation estimation is relaxed to a semidefinite program (SDP). TEASER++ is currently the fastest robust registration algorithm, but similarly, when the number of correspondences is vast, the efficiency of the algorithm decreases significantly and requires a lot of memory size. Due to the points of correspondence can be seen as the nodes of the graph, and the geometric relationship between the correspondences can be represented by the edges of the graph, the outlier removal method combined with graph-theoretic has attracted extensive attention. Clipper \cite{[37]} formulated the outlier removal problem in a graph-theoretic framework using the notion of geometric consistency. It finds the consistent association (inliers) by finding the densest graph and maintains low time complexity through the projection gradient ascent with backtracking line search. Clipper was shown to consistently execute with low runtime and to outperform the state of the art. However, it has not been verified in the registration of real-world data. In addition, the outlier removal strategy based on RANSAC is also widely studied \cite{[38],[39],[40]}. However, the main limitation of these methods is that they need a lot of iterations to find a satisfactory solution under a high outlier rate.

In summary, the main difficulty in the current outlier removal methods is that when the number of correspondence to be processed is large, the algorithms are inefficient. When the outlier rate is very high, it will affect the accuracy of outlier removal and recall rate of inliers, and will also affect the efficiency of the algorithms. Even the current state-of-art methods can not do well in efficiency and accuracy at the same time when there are extremely high outlier rates and a large number of correspondence candidates.

\section{THE THEORY BASIS}\label{sec:the theory basis}

\subsection{A DOF curtailment process based on point-by-point alignment}\label{sec:dof curtailment}
\begin{figure}[!t]
  \centering
  \includegraphics[width=1\columnwidth]{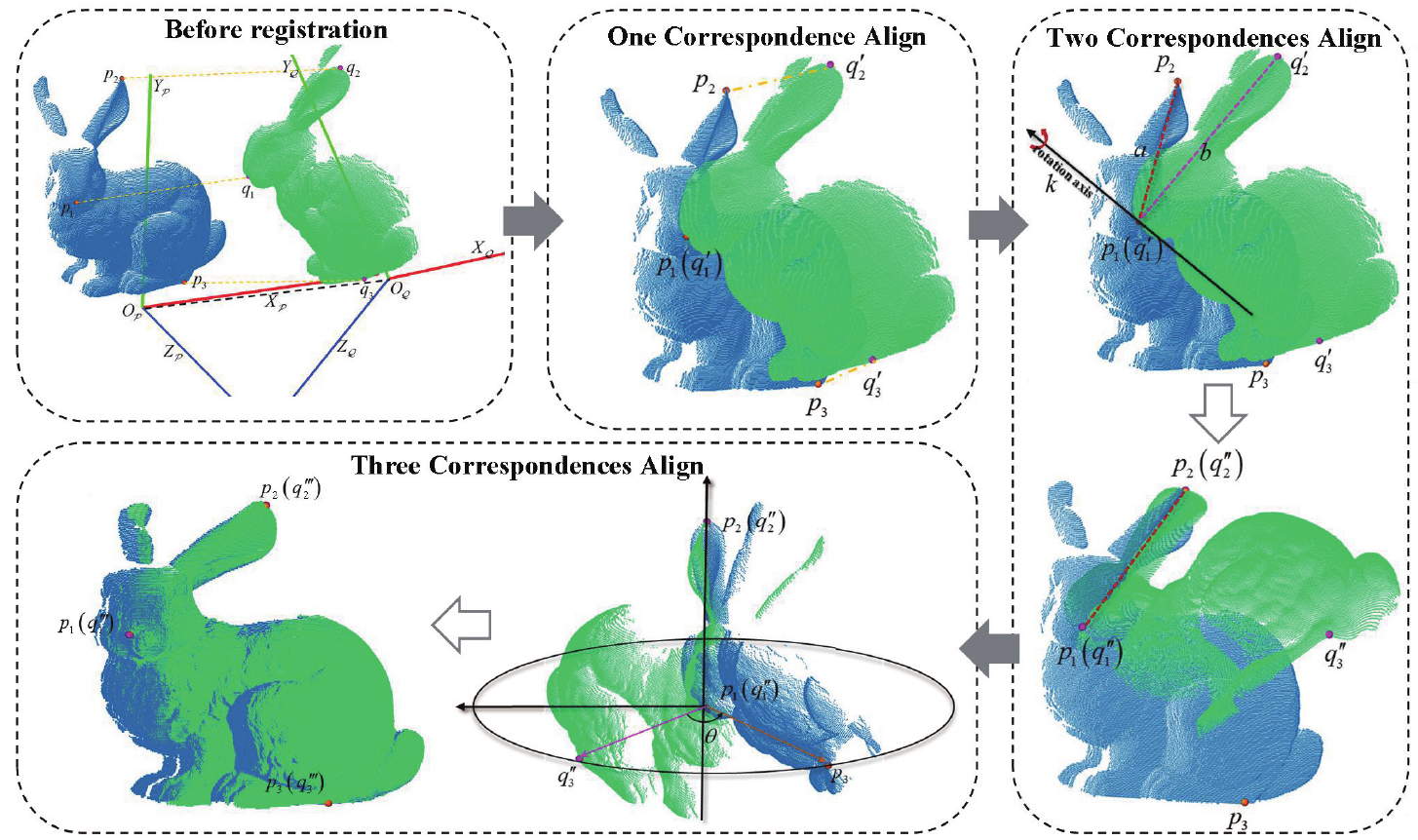}\\
  \caption{The process of three correspondences alignment. }
  \label{Fig1}
\end{figure}

In the Bursa-Wolf model, the angle transformation process is performed as follows: firstly, rotate the coordinate around the $Z-axis$ with an angle ${{\cal{E}}_{Z}}$ to obtain the rotation matrix ${{\mathbf{R}}_{1}}$, then rotate the coordinate system around the new $Y-axis$ with ${{\cal{E}}_{Y}}$ to obtain the matrix ${{\mathbf{R}}_{2}}$, and finally rotate the coordinate system around the new $X-axis$ with ${{\cal{E}}_{X}}$ to get ${{\mathbf{R}}_{3}}$, then the rotation matrix $\mathbf{R}({{\cal{E}}_{X}},{{\cal{E}}_{Y}},{{\cal{E}}_{Z}})={{\mathbf{R}}_{3}}{{\mathbf{R}}_{2}}{{\mathbf{R}}_{1}}$. Different from the coordinate transformation process of the Bursa-Wolf model, in order to combine the coordinate transformation process with the outlier removal to implement an efficient and accurate registration, this paper presents the coordinate transformation process with point-by-point alignment of three correspondences and decouples the registration process into one translation process and two rotation processes around the axis. As shown in Fig. \ref{Fig1}: there are three correspondences without noise and outliers $\left\{ \left( {{\mathbf{p}}_{i}},{{\mathbf{q}}_{i}} \right) \right\}_{i=1}^{3}$ , ${{\mathbf{p}}_{i}},{{\mathbf{q}}_{i}}\in {{\mathbb{R}}^{3}}$. 1) Align ${{\mathbf{q}}_{1}}$ to ${{\mathbf{p}}_{1}}$ is a translation process with the translation matrix obtained as ${{\mathbf{t}}_{1}}={{\mathbf{p}}_{1}}-{{\mathbf{q}}_{1}}$, this step curtails the degree of freedom from 6 to 3; 2) Let $\bf{a}={{\mathbf{p}}_{2}}-{{\mathbf{p}}_{1}}$, $\bf{b}={{\mathbf{{q}'}}_{2}}-{{\mathbf{{q}'}}_{1}}$, then the alignment of ${{\mathbf{{q}'}}_{2}}$ to ${{\mathbf{p}}_{2}}$ can be achieved by rotating vector $\bf b$ around rotation axis ${\bf{k}_{ab}}=\bf a\times \bf b$ by a given angle ${\bf{\theta }_{ab}}=\text{arccos}\left( \bf a\cdot \bf b/\left| \bf a \right|\left| \bf b \right| \right)$. This process fixes two DOF and reduces it from 3 to 1. Let the matrix computed in this step be ${{\mathbf{R}}_{1}}$; 3) The alignment between ${{\mathbf{p}}_{3}}$ and ${{\mathbf{{q}''}}_{3}}$ is also a rotation transformation around the axis. It takes $\bf a$ as the axis of rotation and the angle of rotation is $\theta $ (the method to calculate $\theta $ will be given in Section \ref{sec:Reliability of correspondece} ), this process solves the last one DOF of the registration problem, Let the matrix calculated in this step is ${{\mathbf{R}}_{2}}$. Then the overall rotation matrix $\mathbf{R}={{\mathbf{R}}_{2}}{{\mathbf{R}}_{1}}$.

\subsection{Problem formulation}\label{sec:problem formulation}
The above alignment process is based on the premise that there is no noise and outliers in the correspondences. However, there are a large number of outliers in the set of correspondences which are obtained by feature descriptor matching. Given a correspondence set $\mathcal{H} =\left\{ \left( {{\mathbf{p}}_{i}},{{\mathbf{q}}_{i}} \right) \right\}_{1}^{N}$, the correspondence-based 6-DOF point cloud registration can be formulated as a Truncated Least Squares (TLS) problem \cite{[36]} considering the existence of noise and outliers:
\begin{equation}
	\begin{aligned}
    \mathop {\min }\limits_{{\bf{R}} \in SO(3),{\bf{t}}{ \in \mathbb{R}^3}}\sum\limits_{i = 1}^N {\min \left( {\frac{1}{{\delta _i^2}}{{\left\| {{{\bf{p}}_i} - \left( {{\bf{R}}{{\bf{q}}_i} + {\bf{t}}} \right)} \right\|}^2},{{\overline c }^2}} \right)}
	\end{aligned}
  \label{TLS}
\end{equation}

where $\mathbf{R}\in SO(3)$ is an orthogonal matrix, $\mathbf{t}$ is a $3\times 1$ translation vector, $\delta {}_{i}$ is a noise bound, and ${{\overline{c}}^{2}}$ is a proportional coefficient which can dispose of potential outliers in a rigorous or more tolerant way \cite{[41]} and it is usually set to 1 \cite{[36]}. And for a correspondence:
\begin{equation}
	\begin{cases}
    {\begin{array}{*{20}{l}}
      {\frac{1}{{\delta _i^2}}{{\left\| {{{\bf{p}}_i} - \left( {{\bf{R}}{{\bf{q}}_i} + {\bf{t}}} \right)} \right\|}^2} \le {{\overline c }^2},inlier}\\
      {\frac{1}{{\delta _i^2}}{{\left\| {{{\bf{p}}_i} - \left( {{\bf{R}}{{\bf{q}}_i} + {\bf{t}}} \right)} \right\|}^2} > {{\overline c }^2},outlier}
      \end{array}}
	\end{cases}
  \label{inliers}
\end{equation}

Essentially, TLS estimation is related to Consensus Maximization \cite{[42]}, and the problem (\ref{TLS}) can be reformulated as a maximum consensus problem.
\begin{equation}
	\begin{aligned}
    \begin{array}{l}
      \mathop {\operatorname{maximize}}\limits_{{\bf{R}},{\bf{t}},{\cal I} \subseteq {\cal H}} {\rm{ }}\left| {\cal I} \right|\\
      \textit { subject to }\left\| {{{\bf{p}}_i} - ({\bf{R}}{{\bf{q}}_i} + {\bf{t}})} \right\| < \delta ,{\rm{ }}\forall \left( {{{\bf{p}}_i},{{\bf{q}}_i}} \right) \in {\cal I}
      \end{array}
	\end{aligned}
  \label{maximum consensus}
\end{equation}

where $\left\| \cdot  \right\|$ is defined as the Euclidean distance and the subset $\mathcal{I}$ is often referred to as the consensus set (inliers). The optimal transformation parameter $\widetilde{\mathbf{R}},\widetilde{\mathbf{t}}$ enables (\ref{maximum consensus}) to obtain the maximum consensus set. 

\section{METHODOLOGY}\label{sec:methodology}
\subsection{Preparation}\label{sec:preparation}
Before presenting our method, we first give the approach for obtaining the correspondences: 1) To align the point cloud with a large-scale, the original point cloud is downsampled using the voxel grid method \cite{[43]}, and the grid size $\rho $ is set as the resolution of the new point cloud; 2)The ISS algorithm \cite{[23]} is used to obtain the key points for correspondences matching, let the key point set of the source and target point clouds be $\mathcal{Q}=\{{{\mathbf{q}}_{i}}\}_{i=1}^{n}$ and $\mathcal{P}=\{{{\mathbf{p}}_{j}}\}_{j=1}^{m}$, respec-tively;  3)  Use the FPFH descriptor \cite{[11]} to describe the feature relationship between the key points and its neighbors; 4) Based on the FPFH vector of key points, ${{\mathbf{q}}_{i}}\in \mathcal{Q}$ is used as a query point and ${{\mathbf{p}}_{i}}\in \mathcal{P}$ is obtained by nearest neighbor query using KD-Tree, then the set of correspondences is obtained as $\mathcal{H} =\left\{ ({{\mathbf{p}}_{i}},{{\mathbf{q}}_{i}}) \right\}_{1}^{N}$.

\subsection{An optimal selection strategy based on the reliability of correspondence graph nodes}\label{sec:optimal selection}

An undirected graph can be represented as $\mathcal{G}=(\mathcal{V},\mathcal{E})$, where $\mathcal{V}=\left\{ {{V}_{i}} \right\}_{i=1}^{N}$ is defined as a discrete set of nodes, $\mathcal{E}$ is a set of undirected edges, and $\mathcal{E}\subseteq \mathcal{V}\times \mathcal{V}$ such that $({{V}_{i}},{{V}_{j}})=({{V}_{j}},{{V}_{i}})$. In this paper, the FRGM \cite{[44]} method is used to define the properties of the graph: in Euclidean space, the property of graph nodes is defined as the point coordinate of correspondences i.e $V_{i}^{\mathcal{P}}={{\mathbf{p}}_{i}}$,$V_{i}^{\mathcal{Q}}={{\mathbf{q}}_{i}}$. The edge property of the graph is defined as the length between nodes i.e $\left\| \mathcal{E}_{ij}^{{}} \right\|=\left\| V_{i}^{{}}-V_{j}^{{}} \right\|$, $\left\| \cdot  \right\|$ is Euclidean ${{l}_{2}}-$norm. We use the adjacency matrix $\mathcal{A}$ to denote the adjacency relations between the nodes of a graph $\mathcal{G}=(\mathcal{V},\mathcal{E})$, where $\mathcal{A}$  is a symmetric matrix, if there exists an edge ${{\mathcal{E}}_{ij}}$ between nodes ${{V}_{i}}$ and ${{V}_{j}}$, then ${{a}_{ij}}\in \mathcal{A}=1$, otherwise ${{a}_{ij}}=0$. We establish two undirected graphs  ${{\mathcal{G}}^{\mathcal{P}}}=({{\mathcal{V}}^{\mathcal{P}}},{{\mathcal{E}}^{\mathcal{P}}})$ and ${{\mathcal{G}}^{Q}}=({{\mathcal{V}}^{Q}},{{\mathcal{E}}^{Q}})$ based on correspondences $\mathcal{H} =\left\{ \left( {{\mathbf{p}}_{i}},{{\mathbf{q}}_{i}} \right) \right\}_{1}^{N}$as the correspondence graphs $({{\mathcal{G}}^{\mathcal{P}}},{{\mathcal{G}}^{\mathcal{Q}}})$. First, connect each pair of nodes $(V_{i}^{\mathcal{P}}(V_{i}^{Q}),V_{j}^{\mathcal{P}}(V_{j}^{Q}))$ in ${{\mathcal{G}}^{\mathcal{P}}}({{\mathcal{G}}^{Q}})$ with an edge $\mathcal{E}_{_{ij}}^{\mathcal{P}}(\mathcal{E}_{_{ij}}^{Q})$ to form two complete graphs. Then we have $a_{_{ij}}^{\mathcal{P}}(a_{_{ij}}^{Q})=1,\text{ }a_{_{ii}}^{\mathcal{P}}(a_{_{ii}}^{Q})=0,\text{ }(i=1,2...N,i\ne j)$ in the matrix${{\mathcal{A}}^{\mathcal{P}}}({{\mathcal{A}}^{Q}})$. If $(V_{i}^{\mathcal{P}},V_{i}^{Q})$ and $(V_{j}^{\mathcal{P}},V_{j}^{Q})$ are two sets of correctly matched correspondences, and $\left\| \mathcal{E}_{ij}^{\mathcal{P}} \right\|=\left\| V_{i}^{\mathcal{P}}-V_{j}^{\mathcal{P}} \right\|$, $\left\| \mathcal{E}_{ij}^{\mathcal{Q}} \right\|=\left\| V_{i}^{\mathcal{Q}}-V_{j}^{\mathcal{Q}} \right\|$, then, if $\left\| \mathcal{E}_{ij}^{\mathcal{P}} \right\|-\left\| \mathcal{E}_{ij}^{Q} \right\|=0$, we define $\mathcal{E}_{ij}^{\mathcal{P}}$ and $\mathcal{E}_{ij}^{\mathcal{Q}}$ as correspondence edges. Considering the noise, the constraint condition is $\left\| \mathcal{E}_{ij}^{\mathcal{P}} \right\|-\left\| \mathcal{E}_{ij}^{Q} \right\|<\delta $. Based on this principle, the elements in the adjacency matrix  ${{\mathcal{A}}^{\mathcal{P}}}({{\mathcal{A}}^{Q}})$ that do not satisfy the constraint are reassigned values as:
\begin{equation}
  a_{ij}^{\cal P} = a_{ij}^Q = \left\{
  \begin{aligned}
    {\begin{array}{*{20}{l}}
      {1{\rm{ , }}\left\| {{\cal E}_{ij}^{\cal P}} \right\| - \left\| {{\cal E}_{ij}^Q} \right\| < \delta }\\
      {0{\rm{ , }}\left\| {{\cal E}_{ij}^{\cal P}} \right\| - \left\| {{\cal E}_{ij}^Q} \right\| \ge \delta }
      \end{array}} {\rm{ , }}i = 1,2,...,N{\rm{ }}i \ne j
  \end{aligned}
  \right.
  \label{reassigned values}
\end{equation}

Since ${{\mathcal{A}}^{\mathcal{P}}}\equiv {{\mathcal{A}}^{Q}}$, we directly use ${{\mathcal{A}}^{\mathcal{P}\mathcal{Q}}}$ to denote these two identical adjacency matrices.

\textbf{Definition 1}:  \textit{The sum of the elements of the $i$-$st$ row (column) in $\mathcal{A}$ is equal to the degree of vertex (node) in a graph, denoted as:}
\begin{equation}
  \begin{aligned}
    {{\cal D}_{\cal G}}\left( {{V_i}} \right) = \sum\limits_{j = 1}^N {a_{ij}^{}{\rm{ }}\left( {a_{ij}^{} \in {{\cal A}^{}}} \right)} 
  \end{aligned}
  \label{node reliability}
\end{equation}

\textbf{Proposition 1}:  \textit{The matching reliability of the correspondence $(V_{i}^{\mathcal{P}},V_{i}^{Q})$ can be measured by the degree ${{\mathcal{D}}_{\mathcal{G}}}({{V}_{i}})$ of the adjacency matrix $\mathcal{A}$, The larger the reliability of a node ${{V}_{i}}$ represents that the node satisfies more constraints between corresponding graphs $({{\mathcal{G}}^{\mathcal{P}}},{{\mathcal{G}}^{\mathcal{Q}}})$, and the higher the matching reliability it has.}

\begin{figure}[!t]
  \centering
  \includegraphics[width=1\columnwidth]{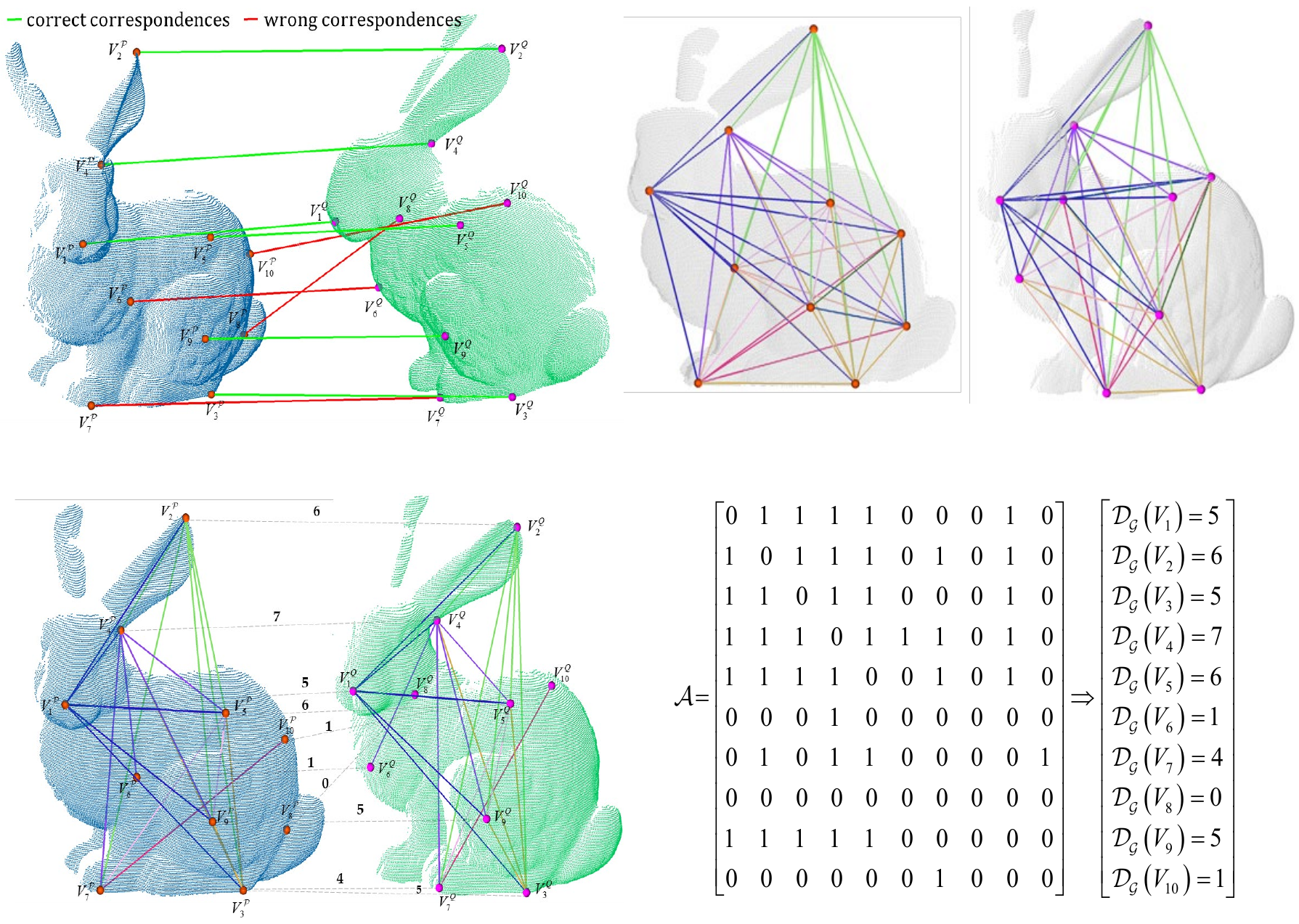}\\
  \caption{An example for computing the node reliability metric. (a) Ten correspondences for example, (b) undirected complete graph, (c) compact graph, and (d) adjacency matrix and degree of nodes. }
  \label{Fig2}
\end{figure}

Fig. \ref{Fig2} shows an example, there are 10 sets of correspondence: $\mathcal{H}=\left\{ ({{\mathbf{p}}_{i}},{{\mathbf{q}}_{i}}) \right\}_{1}^{10}$, where 6,7,8,10 are outlier correspondences as shown in Fig. \ref{Fig2}(a). Firstly, two undirected complete topological graph is constructed based on the correspondences as in Fig. \ref{Fig2}(b), then the adjacency matrix $\mathcal{A}$ according to the constraint of correspondence edges is reassigned using (\ref{reassigned values}). Fig. \ref{Fig2}(c) shows a compact graph retaining the edges that satisfy the constraint, and Fig. \ref{Fig2}(d) shows the corresponding adjacency matrix $\mathcal{A}$  of the compact graph and the degree matrix $\mathcal{D}$ for each node. In $\mathcal{D}$, the reliability degrees of the outlier nodes 6,7,8,10 are 1,4,0,1, respectively, and their degrees rank last among all nodes and have the lowest matching reliability.

The point cloud registration requires at least three sets of correspondence to complete the fixation of 6 degrees of freedom. Due to the existence of noise and outlier points, more correspondences are used to obtain the optimal registration accuracy, so selecting a subset of correspondences with high reliability can effectively reduce the influence of outliers and the number of correspondences to be checked, thus improving the efficiency of the algorithm. According to the above principle, we sort all the elements in the matrix $\cal D$ and then select the top $K$ sets of reliable point pairs to form a reliable set of correspondences $\mathcal{H}{{}_{r}} =\left\{ ({{\mathbf{p}}_{i}},{{\mathbf{q}}_{i}}) \right\}_{1}^{K}$, $\mathcal{H}{{}_{r}}\subseteq \mathcal{H}$, and proceed to the next alignment step.

\subsection{Alignment based on point-by-point method with edge reliability}\label{sec:edge reliability}
\subsubsection{Two point pairs alignment based on the correspondence graph}\label{sec:two point pairs alignment}

\begin{figure}[!t]
    \centering
    \includegraphics[width=1\columnwidth]{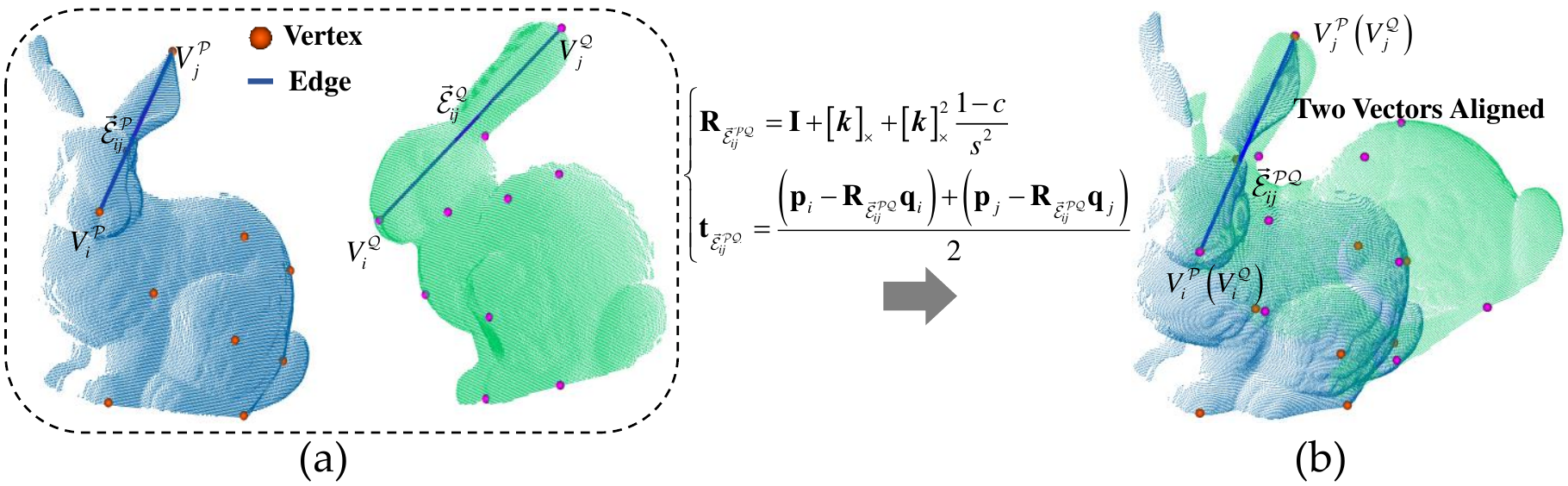}\\
    \caption{Two point pairs (correspondences) aligning process based on the correspondence graph. (a) Before aligning, and (b) after aligning. }
    \label{Fig3}
  \end{figure}

Based on the reliable correspondence set $\mathcal{H}{{}_{r}} =\left\{ ({{\mathbf{p}}_{i}},{{\mathbf{q}}_{i}}) \right\}_{1}^{K}$ we reconstruct two undirected graphs  ${{\mathcal{G}}^{\mathcal{P}}}=({{\mathcal{V}}^{\mathcal{P}}},{{\mathcal{E}}^{\mathcal{P}}})$ and ${{\mathcal{G}}^{Q}}=({{\mathcal{V}}^{Q}},{{\mathcal{E}}^{Q}})$ whose edges satisfy the constraint $\left\| \mathcal{E}_{ij}^{\mathcal{P}} \right\|-\left\| \mathcal{E}_{ij}^{Q} \right\|<\delta$. As shown in Fig. \ref{Fig3}(a): we choose a correspondence edge $(\mathcal{E}_{ij}^{\mathcal{P}},\mathcal{E}_{ij}^{Q})$ from ${{\mathcal{G}}^{\mathcal{P}}}$ and ${{\mathcal{G}}^{Q}}$, where the vertices of $\mathcal{E}_{ij}^{\mathcal{P}}$ are $V_{i}^{\mathcal{P}}$ and $V_{j}^{\mathcal{P}}$, the vertices of $\mathcal{E}_{ij}^{\mathcal{P}}$ are $V_{i}^{Q}$ and $V_{j}^{Q}$. Following the point-by-point alignment method described in Section \ref{sec:dof curtailment}, we should first perform a translation process to align $V_{i}^{\mathcal{P}}(V_{j}^{\mathcal{P}})$ and $V_{i}^{\mathcal{Q}}(V_{j}^{\mathcal{Q}})$, and then align $V_{j}^{\mathcal{P}}(V_{i}^{\mathcal{P}})$ and $V_{i}^{\mathcal{Q}}(V_{j}^{\mathcal{Q}})$ by a rotation process around the axis. In practice, two edges $\vec{\mathcal{E}}_{ij}^{\mathcal{P}}=\overrightarrow{V_{i}^{\mathcal{P}}V_{j}^{\mathcal{P}}}={{\mathbf{p}}_{j}}-{{\mathbf{p}}_{i}}$, $\vec{\mathcal{E}}_{ij}^{\mathcal{Q}}=\overrightarrow{V_{i}^{\mathcal{Q}}V_{j}^{\mathcal{Q}}}={{\mathbf{q}}_{j}}-{{\mathbf{q}}_{i}}$ can be represented by two vectors, and then the process of 1) 2) in the point-by-point alignment can be combined into the process of aligning the two vectors. Let $\textbf{\textit{k}}=(\vec{\mathcal{E}}_{ij}^{\mathcal{P}}/\left\| \vec{\mathcal{E}}_{ij}^{\mathcal{P}} \right\|)\times (\vec{\mathcal{E}}_{ij}^{Q}/\left\| \vec{\mathcal{E}}_{ij}^{Q} \right\|)$, $s=\left\| \textbf{\textit{k}} \right\|$ (sine of angle), $c=\vec{\mathcal{E}}_{ij}^{\mathcal{P}}\cdot \vec{\mathcal{E}}_{ij}^{Q}$ (cosine of angle). Then the rotation matrix ${{\mathbf{R}}_{{{{\vec{\mathcal{E}}}}_{ij}}}}$ and translation matrix ${{\mathbf{t}}_{\vec{\mathcal{E}}_{ij}^{\mathcal{P}\mathcal{Q}}}}$ between vectors $\vec{\mathcal{E}}_{ij}^{\mathcal{Q}}$ and $\vec{\mathcal{E}}_{ij}^{\mathcal{P}}$ can be calculated by the following equation:
\begin{equation}
  \left\{
  \begin{aligned}
    {\begin{array}{*{35}{l}}
       {{\mathbf{R}}_{\vec{\mathcal{E}}_{ij}^{\mathcal{P}\mathcal{Q}}}}=\mathbf{I}+{{\left[ k \right]}_{\times }}+\left[ k \right]_{\times }^{2}\frac{1-c}{{{s}^{2}}}  \\
   {{\mathbf{t}}_{\vec{\mathcal{E}}_{ij}^{\mathcal{P}\mathcal{Q}}}}=\frac{\left( {{\mathbf{p}}_{i}}-{{\mathbf{R}}_{\vec{\mathcal{E}}_{ij}^{\mathcal{P}\mathcal{Q}}}}{{\mathbf{q}}_{i}} \right)+\left( {{\mathbf{p}}_{j}}-{{\mathbf{R}}_{\vec{\mathcal{E}}_{ij}^{\mathcal{P}\mathcal{Q}}}}{{\mathbf{q}}_{j}} \right)}{2}  \\
      \end{array}}
  \end{aligned}
  \right.
  \label{two edge alignment}
\end{equation}

Where $\vec{\mathcal{E}}_{ij}^{\mathcal{P}\mathcal{Q}}$ denotes the aligned edge vector and ${{\left[ \textbf{\textit{k}} \right]}_{\times }}$ is the skew-symmetric cross-product matrix of $\textbf{\textit{k}}$ \cite{[45]} :
\begin{equation}
  \begin{aligned}
	  {{\left[ \textbf{\textit{k}} \right]}_{\times }}\triangleq \left[ \begin{matrix}
    0 & -{{\textbf{\textit{k}}}_{3}} & {{\textbf{\textit{k}}}_{2}}  \\
    {{\textbf{\textit{k}}}_{3}} & 0 & -{{\textbf{\textit{k}}}_{1}}  \\
    -{{\textbf{\textit{k}}}_{2}} & {{\textbf{\textit{k}}}_{1}} & 0  \\
\end{matrix} \right]
  \end{aligned}
  \label{matrix k}
\end{equation}
The last part of the formula for ${{\mathbf{R}}_{\vec{\mathcal{E}}_{ij}^{\mathcal{P}\mathcal{Q}}}}$ can be simplified as :
\begin{equation}
  \begin{aligned}
    \frac{1-c}{{{s}^{2}}}=\frac{1-c}{1-{{c}^{2}}}=\frac{1}{1+c}
  \end{aligned}
  \label{R simplified}
\end{equation}
  ${{\mathbf{t}}_{\vec{\mathcal{E}}_{ij}^{\mathcal{P}\mathcal{Q}}}}$ denotes the average translation vector of two correspondences $({{\mathbf{p}}_{i}},{{\mathbf{q}}_{i}})$ and $\left( {{\mathbf{p}}_{j}},{{\mathbf{q}}_{j}} \right)$ after alignment. All the source points are transformed using transformation parameters ${{\mathbf{R}}_{\vec{\mathcal{E}}_{ij}^{\mathcal{P}\mathcal{Q}}}}$ and ${{\mathbf{t}}_{\vec{\mathcal{E}}_{ij}^{\mathcal{P}\mathcal{Q}}}}$, then a new set of correspondence $\mathcal{H}{{{}'}_{r}}=\left\{ ({{\mathbf{p}}_{i}},{{{\mathbf{{q}'}}}_{i}})\left| {{{\mathbf{{q}'}}}_{i}}={{\mathbf{R}}_{\vec{\mathcal{E}}_{ij}^{\mathcal{P}\mathcal{Q}}}}{{\mathbf{q}}_{i}}+{{\mathbf{t}}_{\vec{\mathcal{E}}_{ij}^{\mathcal{P}\mathcal{Q}}}} \right. \right\}_{1}^{K}$can be obtained. The result after alignment is shown in Fig. \ref{Fig3}(b).
  \subsubsection{Reliability of the correspondence edge after alignment}\label{sec:Reliability of correspondece}
  The above procedure combines the 1) and 2) steps in Section \ref{sec:dof curtailment} to fix 5-DOF in the registration problem, leaving only the last one DOF to be solved. After the two correspondences $({{\mathbf{p}}_{i}},{{\mathbf{q}}_{i}})$ and $({{\mathbf{p}}_{j}},{{\mathbf{q}}_{j}})$ are aligned, there are still $K-2$ correspondences in ${{{\mathcal{H}}'}_{r}}$ have not been aligned. Let $(V_{k}^{\mathcal{P}}\in {{\mathcal{G}}^{\mathcal{P}}},V_{k}^{\mathcal{Q}}\in {{\mathcal{G}}^{\mathcal{Q}}})\in {{{\mathcal{H}}'}_{r}}$, $k=1,2,...,K;\text{ }k\ne i\ne j$ is a correspondence that is unaligned. We define an Edge-Node Affinity Matrix (ENAM) $\mathcal{M}$ (the size is $\left( K-2 \right)\times 1$) to evaluate the reliability of a correspondence edge after alignment. The values of the elements in the matrix $\mathcal{M}$ of $\mathcal{E}_{ij}^{\mathcal{P}\mathcal{Q}}$ can be obtained as follows:
\begin{equation}
    \begin{aligned}
    m(k,1) = \left\{
    {\begin{array}{*{35}{l}}
        {1{\rm{ , }}{\cal F}\left( {{\cal E}_{ij}^{{\cal P}{\cal Q}},{\cal V}_k^{{\cal P}{\cal Q}}} \right) < 0}\\
        {0{\rm{ , }}{\cal F}\left( {{\cal E}_{ij}^{{\cal P}{\cal Q}},{\cal V}_k^{{\cal P}{\cal Q}}} \right) \ge 0}
      \end{array}
    }
  \right.,(i,j,k = 1,2,...,K,i \ne j \ne k)
  \end{aligned}
  \label{matrix M}
\end{equation}

Where,  $\mathcal{E}_{ij}^{\mathcal{P}\mathcal{Q}}$=$(\mathcal{E}_{ij}^{\mathcal{P}},\mathcal{E}_{ij}^{\mathcal{Q}})$, $\mathcal{V}_{k}^{\mathcal{P}\mathcal{Q}}$=$(V_{k}^{\mathcal{P}},V_{k}^{\mathcal{Q}})$,   $\mathcal{F}(\mathcal{E}_{ij}^{\mathcal{P}\mathcal{Q}},\mathcal{V}_{k}^{\mathcal{P}\mathcal{Q}})$ denotes the constraint function satisfied by the aligned edge $\mathcal{E}_{ij}^{\mathcal{P}\mathcal{Q}}$ and the unaligned correspondence node $(V_{k}^{\mathcal{P}},V_{k}^{\mathcal{Q}})$ in the three-dimensional space ${{\mathbb{R}}^{3}}$.

\textbf{Definition 2}:  \textit{The sum of the elements in the Edge-Node Affinity Matrix $\mathcal{M}$ is equal to the degree of aligned edge $\mathcal{E}_{ij}^{{}}$, denoted as:}
\begin{equation}
  \begin{aligned}
    {{\cal D}_{\cal G}}\left( {{{\cal E}_{ij}}} \right) = \sum\limits_{k = 1}^N {m(k,1){\rm{ }}\left( {m(k,1) \in {\cal M}} \right)}
  \end{aligned}
  \label{degree of edge}
\end{equation}

\textbf{Proposition 2}:  \textit{The reliability of the aligned correspondence edge $(\mathcal{E}_{ij}^{\mathcal{P}},\mathcal{E}_{ij}^{Q})$ in the same constraint function can be measured by the degree ${{\mathcal{D}}_{\mathcal{G}}}(\mathcal{E}_{ij}^{\mathcal{P}\mathcal{Q}})$ of the Edge-Node Affinity Matrix $\mathcal{M}$. The greater the reliability of the aligned edge $\mathcal{E}_{ij}^{\mathcal{P}\mathcal{Q}}$, the more correspondence nodes satisfy the constrain function $\mathcal{F}(\mathcal{E}_{ij}^{\mathcal{P}\mathcal{Q}},\mathcal{V}_{k}^{\mathcal{P}\mathcal{Q}})$ after the correspondence $(\mathcal{E}_{ij}^{\mathcal{P}},\mathcal{E}_{ij}^{Q})$ aligned.}

The reliability of the aligned correspondence edge $\mathcal{E}_{ij}^{\mathcal{P}\mathcal{Q}}$ differs depending on the definition of the constraint function $\mathcal{F}(\mathcal{E}_{ij}^{\mathcal{P}\mathcal{Q}},\mathcal{V}_{k}^{\mathcal{P}\mathcal{Q}})$. Here, we give two constraint functions ${{\mathcal{F}}_{1}}(\mathcal{E}_{ij}^{\mathcal{P}\mathcal{Q}},\mathcal{V}_{k}^{\mathcal{P}\mathcal{Q}})$ and ${{\mathcal{F}}_{2}}(\mathcal{E}_{ij}^{\mathcal{P}\mathcal{Q}},\mathcal{V}_{k}^{\mathcal{P}\mathcal{Q}})$ to compute the reliability of aligned correspondence edge $\mathcal{E}_{ij}^{\mathcal{P}\mathcal{Q}}$. Where, ${{\mathcal{F}}_{1}}(\mathcal{E}_{ij}^{\mathcal{P}\mathcal{Q}},\mathcal{V}_{k}^{\mathcal{P}\mathcal{Q}})$ is a loose constraint function in the alignment problem, while ${{\mathcal{F}}_{2}}(\mathcal{E}_{ij}^{\mathcal{P}\mathcal{Q}},\mathcal{V}_{k}^{\mathcal{P}\mathcal{Q}})$ is a tight constraint function.

\textbf{A. Loose function ${{\cal F}_1}({\cal E}_{ij}^{{\cal P}{\cal Q}},{\cal V}_k^{{\cal P}{\cal Q}})$  constraint by the node-edge projection distance.}

\begin{figure}[!t]
  \centering
  \includegraphics[width=1\columnwidth]{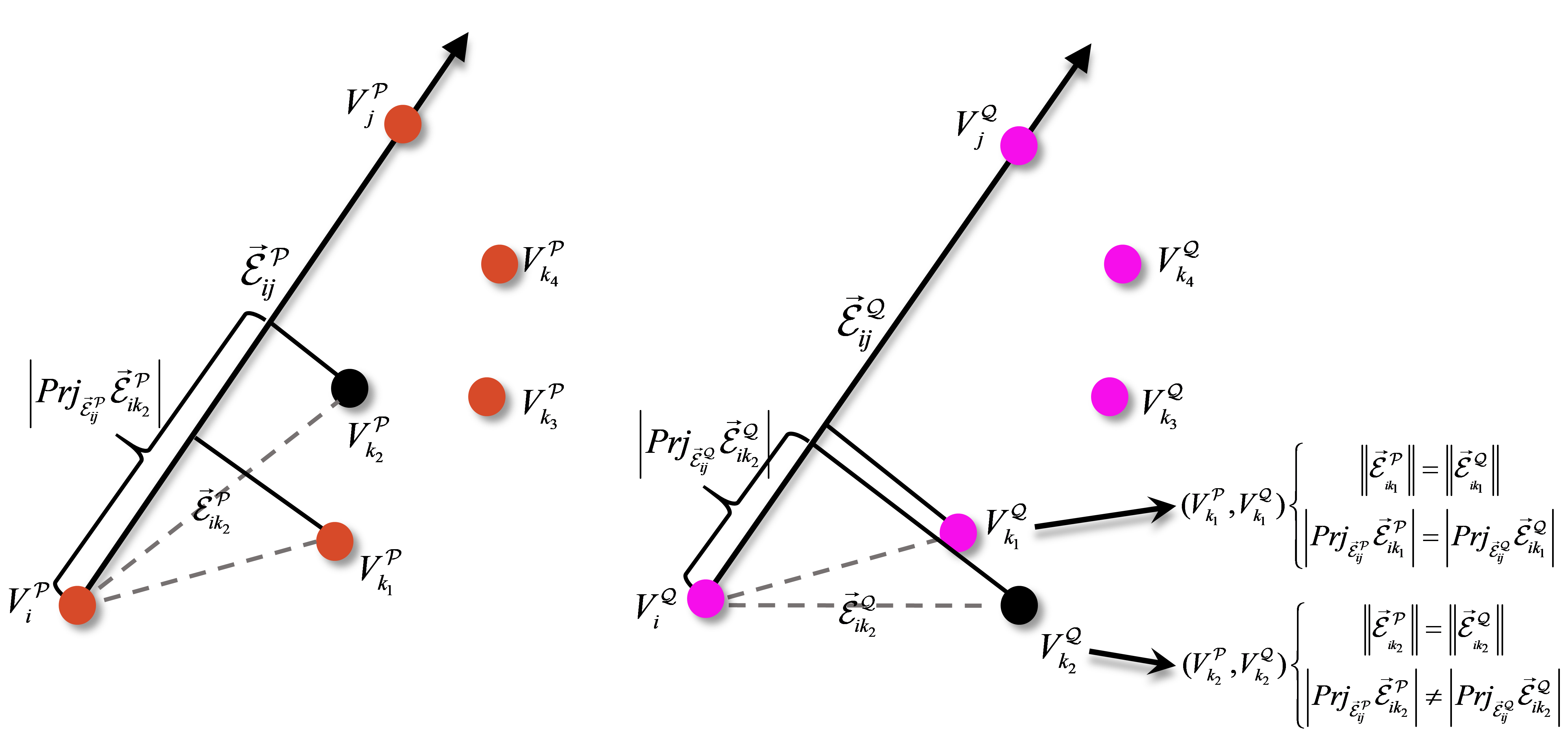}\\
  \caption{An example to explain that the constraint$\left\| \mathcal{E}_{ij}^{\mathcal{P}} \right\|=\left\| \mathcal{E}_{ij}^{Q} \right\|$ does not completely remove outliers. The orange and pink points in the figure represent inliers, and the black points represent outliers. All correspondence satisfies the constraint: $\left\| \vec{\mathcal{E}}_{_{ik}}^{\mathcal{P}} \right\|=\left\| \vec{\mathcal{E}}_{_{ik}}^{\mathcal{Q}} \right\|$ between the edges, but the outliers cannot satisfy the constraint: $\left| Pr{{j}_{\vec{\mathcal{E}}_{ij}^{\mathcal{P}}}}\vec{\mathcal{E}}_{ik}^{\mathcal{P}} \right|=\left| Pr{{j}_{\vec{\mathcal{E}}_{ij}^{\mathcal{Q}}}}\vec{\mathcal{E}}_{ik}^{\mathcal{Q}} \right|$. }
  \label{Fig4}
\end{figure}

In Section \ref{sec:optimal selection} the length error of the correspondence edges is employed to select reliable correspondences, but this still does not completely reject outliers, as an example shown in Fig. \ref{Fig4}.  A loose constraint function ${{\mathcal{F}}_{1}}(\mathcal{E}_{ij}^{\mathcal{P}\mathcal{Q}},\mathcal{V}_{k}^{\mathcal{P}\mathcal{Q}})$ describing the geometric constraint between the unaligned correspondence nodes and the aligned correspondence edge is defined which can further reject some remaining outliers, and evaluate the reliability of the aligned edge under this constraint. Let $(\vec{\mathcal{E}}_{ij}^{\mathcal{P}}=\overrightarrow{V_{i}^{\mathcal{P}}V_{j}^{\mathcal{P}}},\vec{\mathcal{E}}_{ij}^{Q}=\overrightarrow{V_{i}^{\mathcal{Q}}V_{j}^{\mathcal{Q}}})$ is the aligned correspondence edge vector can be represented as $\vec{\mathcal{E}}_{ij}^{\mathcal{P}\mathcal{Q}}$, $(V_{k}^{\mathcal{P}},V_{k}^{\mathcal{Q}})$ represents an unaligned correspondence node. Constructs the vector edge $\vec{\mathcal{E}}_{ik}^{\mathcal{P}}=\overrightarrow{V_{i}^{\mathcal{P}}V_{k}^{\mathcal{P}}},\vec{\mathcal{E}}_{ik}^{Q}=\overrightarrow{V_{i}^{\mathcal{Q}}V_{k}^{\mathcal{Q}}}$ then we use the projection distance of the edge vector $\vec{\mathcal{E}}_{ik}^{\mathcal{P}}(\vec{\mathcal{E}}_{ik}^{Q})$ to $\vec{\mathcal{E}}_{ij}^{\mathcal{P}}(\vec{\mathcal{E}}_{ij}^{\mathcal{Q}})$ as a constraint, and ${{\mathcal{F}}_{1}}(\mathcal{E}_{ij}^{\mathcal{P}\mathcal{Q}},\mathcal{V}_{k}^{\mathcal{P}\mathcal{Q}})$ can be expressed as:
\begin{equation}
  \begin{aligned}
    {{\cal F}_1}({\cal E}_{ij}^{{\cal P}{\cal Q}},{\cal V}_k^{{\cal P}{\cal Q}}) = \left| {\left| {Pr{j_{\vec {\cal E}_{ij}^{\cal P}}}\vec {\cal E}_{ik}^{\cal P}} \right| - \left| {Pr{j_{\vec {\cal E}_{ij}^{\cal Q}}}\vec {\cal E}_{ik}^{\cal Q}} \right|} \right| - \delta
  \end{aligned}
  \label{loose function}
\end{equation}

Where $Pr{{j}_{{{{\vec{\mathcal{E}}}}_{1}}}}{{\vec{\mathcal{E}}}_{2}}$ denotes the projection distance from ${{\vec{\mathcal{E}}}_{2}}$ to ${{\vec{\mathcal{E}}}_{1}}$ and $\delta $ is the distance tolerance threshold. All points in $\mathcal{H}{{{}'}_{r}}$ are checked for the above constraints, and the edge-node association matrix ${{\mathcal{M}}_{1}}$ is calculated according to (\ref{matrix M}) then the reliability of aligned edge $\mathcal{E}_{ij}^{\mathcal{P}\mathcal{Q}}$ under ${{\mathcal{F}}_{1}}(\mathcal{E}_{ij}^{\mathcal{P}\mathcal{Q}},\mathcal{V}_{k}^{\mathcal{P}\mathcal{Q}})$ constraint is ${{\mathcal{D}}_{\mathcal{G}}}(\mathcal{E}_{ij}^{\mathcal{P}\mathcal{Q}})\left| _{{{\mathcal{F}}_{1}}} \right.$.

Although the constraint function ${{\mathcal{F}}_{1}}(\mathcal{E}_{ij}^{\mathcal{P}\mathcal{Q}},\mathcal{V}_{k}^{\mathcal{P}\mathcal{Q}})$ can further reject some outliers, it is not able to complete the alignment of the point cloud under this constraint, so it is a loose constraint function.

\textbf{B. Tight function ${{\cal F}_2}({\cal E}_{ij}^{{\cal P}{\cal Q}},{\cal V}_k^{{\cal P}{\cal Q}})$  to complete the registration.}

If two point clouds can be aligned according to a constraint, then this constraint is a tight constraint for the registration problem. We define the tight constraint function as:
\begin{equation}
  \begin{aligned}
    {{\cal F}_2}({\cal E}_{ij}^{{\cal P}{\cal Q}},{\cal V}_k^{{\cal P}{\cal Q}}) = \left\| {V_k^{\cal P} - {\bf{R}}(\theta ,\vec {\cal E}_{ij}^{{\cal P}{\cal Q}})V_k^{\cal Q}} \right\| - \delta 
  \end{aligned}
  \label{tight function}
\end{equation}

The (\ref{tight function}) indicates that the aligned edge vector $\vec{\mathcal{E}}_{ij}^{\mathcal{P}\mathcal{Q}}$ is used as the rotation axis and the rotation angle $\theta $ aligns the third correspondence $(V_{k}^{\mathcal{P}},V_{k}^{\mathcal{Q}})$, This completes the reduction of the last 1-DOF and finishes the 6-DOF curtailment problem based on three correspondences (step 3) in Section \ref{sec:dof curtailment}). $\mathbf{R}(\theta ,\vec{\mathcal{E}}_{ij}^{\mathcal{P}\mathcal{Q}})$ represents the rotation matrix obtained by rotating the angle $\theta $ with $\vec{\mathcal{E}}_{ij}^{\mathcal{P}\mathcal{Q}}$ as the axis. Now, we introduce the method to calculate the angle $\theta $.

\begin{figure}[!t]
  \centering
  \includegraphics[width=1\columnwidth]{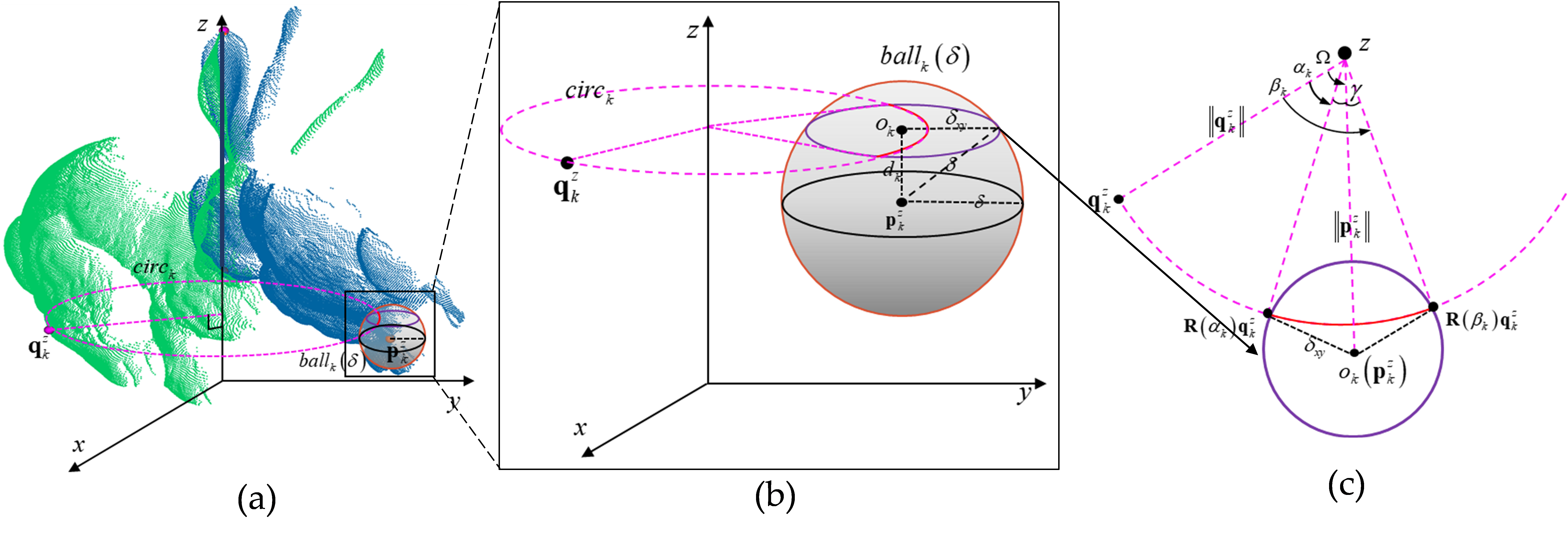}\\
  \caption{Calculation method of $\theta $. (a) Rotate the vector $\vec{\mathcal{E}}_{ij}^{\mathcal{P}\mathcal{Q}}$ to the Z-axis direction, and the correspondence to be aligned is $\left( \mathbf{p}_{k}^{z},\mathbf{q}_{k}^{z} \right)$, (b) calculate the angle interval that aligns $\left( \mathbf{p}_{k}^{z},\mathbf{q}_{k}^{z} \right)$, ${{d}_{k}}=\left| \mathbf{q}_{k}^{z}(3)-\mathbf{p}_{k}^{z}(3) \right|$,${{\delta }_{xy}}=\sqrt{{{\delta }^{2}}-d_{k}^{2}}$, and (c) the top view of (b), $\Omega =azi(\mathbf{p}_{k}^{z})-azi(\mathbf{q}_{k}^{z})$, $azi(\cdot )$is the coordinate azimuth,    $\gamma =\arccos (\left\| \mathbf{q}_{k}^{z} \right\|_{xy}^{2}+\left\| \mathbf{p}_{k}^{z} \right\|_{xy}^{2}-{{\delta }_{xy}})/2{{\left\| \mathbf{q}_{k}^{z} \right\|}_{xy}}{{\left\| \mathbf{p}_{k}^{z} \right\|}_{xy}}$,${{\alpha }_{k}}=\Omega -\gamma $,${{\beta }_{k}}=\Omega +\gamma $. }
  \label{Fig5}
\end{figure}

Assuming that $({{\mathbf{p}}_{k}},{{\mathbf{{q}'}}_{k}})$ is an inlier, we first compute the rotation matrix ${{\mathbf{R}}_{Z}}$ according to (\ref{two edge alignment}), align the vector $\vec{\mathcal{E}}_{ij}^{\mathcal{P}\mathcal{Q}}$ to $Z=\left( 0,0,1 \right)$, let $\mathbf{p}_{k}^{z}={{\mathbf{R}}_{Z}}{{\mathbf{p}}_{k}}$, $\mathbf{q}_{k}^{z}={{\mathbf{R}}_{Z}}{{\mathbf{{q}'}}_{k}}$. As shown in Fig. \ref{Fig5}(a), the alignment between $\mathbf{p}_{k}^{z}$ and $\mathbf{q}_{k}^{z}$ can be regarded as a rotation process of $\mathbf{p}_{k}^{z}$ around the Z-axis by an angle $\theta $. Considering noise it can be expressed as:
\begin{equation}
  \begin{aligned}
    \left\| {{\bf{p}}_k^z - {\bf{R}}(\theta ,Z){\bf{q}}_k^z} \right\| < \delta 
  \end{aligned}
  \label{rotation error}
\end{equation}

Where,$\mathbf{R}(\theta ,Z)$ is the rotation matrix corresponding to the rotation angle $\theta $ with Z-axis as the rotation axis. We refer to the work of Cai et al.\cite{[35]} to represent the alignment error between $\mathbf{p}_{k}^{z}$ and $\mathbf{q}_{k}^{z}$ in 3D space as a ball with $\mathbf{p}_{k}^{z}$ is the center and $\delta $ is the radius:
\begin{equation}
  \begin{aligned}
bal{l_k}\left( \delta  \right) = \left\{ {{\bf{p}}_k^z \in {^3}\left| {{\rm{ }}\left\| {{\bf{p}}_k^z - {\bf{q}}_k^z} \right\| < \delta } \right.} \right\}
  \end{aligned}
  \label{ball error}
\end{equation}

The trajectory of ${\bf{R}}\left( {\theta ,Z} \right){\bf{q}}_k^z$  is a circle, which is denoted as:
\begin{equation}
  \begin{aligned}
    cir{c_k}\left( {{\bf{q}}_k^z} \right) = \left\{ {{\bf{R}}(\theta ,Z){\bf{q}}_k^z\left| {\theta  \in \left[ {0,2\pi } \right]} \right.} \right\}
  \end{aligned}
  \label{circal}
\end{equation}

It is obvious that $(\mathbf{p}_{k}^{z},\mathbf{q}_{k}^{z})$ is aligned by the rotation matrix $\mathbf{R}(\theta ,Z)$ only when $cir{{c}_{k}}$ and $bal{{l}_{k}}\left( \delta  \right)$ intersect (red part of Fig. \ref{Fig5}). As shown in Fig. \ref{Fig5}(c), suppose $cir{{c}_{k}}$ and $bal{{l}_{k}}\left( \delta  \right)$ intersect at points $\mathbf{R}({{\alpha }_{k}})\mathbf{q}_{k}^{z}$ and  $\mathbf{R}\left( {{\beta }_{k}} \right)\mathbf{q}_{k}^{z}$ which are rotated by the angles ${{\alpha }_{k}}$ and ${{\beta }_{k}}$, respectively. Then:
\begin{equation}
  \begin{aligned}
    \theta  \in \left[ {{\alpha _k},{\beta _k}} \right] \subseteq \left[ {0,2\pi } \right]
  \end{aligned}
  \label{theta}
\end{equation}

Where ${\alpha _k}$ and ${{\beta }_{k}}$ are calculated as shown in Fig. \ref{Fig5}(b)(c).

According to the above approach, taking $\vec{\mathcal{E}}_{ij}^{\mathcal{P}\mathcal{Q}}$ as the reference edge, $\forall \left( {{\mathbf{p}}_{k}},{{{\mathbf{{q}'}}}_{k}} \right)\in \mathcal{H}{{{}'}_{r}} (k\ne i\ne j)$ can calculate an angle interval $\theta \in \left[ {{\alpha }_{k}},{{\beta }_{k}} \right]$, and there are $K-2$ in total, However, in $\mathcal{H}{{{}'}_{r}}$, there are both inliers and outliers that are not removed in Section \ref{sec:optimal selection}. So obtaining an optimal $\theta $ can be defined as the problem of finding the maximum consensus set of $\theta $:
\begin{equation}
  \begin{aligned}
  {O_{ij}} = \mathop {\max }\limits_\theta  \sum\limits_{k = 1}^K {\mathbb{I}{_{ij}}} \left( {\theta  \in \left[ {{\alpha _k},{\beta _k}} \right]} \right),(k \ne i \ne j)
  \end{aligned}
  \label{}
\end{equation}

Where ${{O}_{ij}}$ is defined as the maximum number of correspondences that can be aligned by the same rotation angle $\theta $ in the premise that edges $\vec{\mathcal{E}}_{ij}^{\mathcal{P}}$ and $\vec{\mathcal{E}}_{ij}^{Q}$ are aligned. ${{\mathbb{I}}_{ij}}(\cdot )$ is an indicator function with values 0, 1. This is a classical interval stabbing problem in computational geometry \cite{[46]}, and the detailed solution can be found in \cite{[33],[35]}, whose algorithmic time complexity is $\mathcal{O}\left( K\log K \right)$.

We assign the edge-node affinity matrix ${{\mathcal{M}}_{\text{ 2}}}$ based on the value of ${{\mathbb{I}}_{ij}}(\cdot )$. The reliability of aligned correspondence edge $\mathcal{E}_{ij}^{\mathcal{P}\mathcal{Q}}$ subject to the function constraint ${{\mathcal{F}}_{2}}(\mathcal{E}_{ij}^{\mathcal{P}\mathcal{Q}},\mathcal{V}_{k}^{\mathcal{P}\mathcal{Q}})$ is ${{\mathcal{D}}_{\mathcal{G}}}(\mathcal{E}_{ij}^{\mathcal{P}\mathcal{Q}})\left| _{{{\mathcal{F}}_{2}}} \right.$. The maximum consensus set of correspondence computed according to this edge is ${{\mathcal{I}}_{ij}}={{\mathbb{I}}_{ij}}\left( \mathcal{H}{{{{}'}}_{r}} \right)\subseteq \mathcal{H}$.

The final point cloud registration parameters obtained based on the correspondence edge $\left( \mathcal{E}_{ij}^{\mathcal{P}},\mathcal{E}_{ij}^{Q} \right)$ are:
\begin{equation}
  \begin{aligned}
    \left\{ {\begin{array}{*{20}{l}}
      {{{\bf{R}}_{ij}} = {\bf{R}}(\theta ,{{\vec {\cal E}}_{ij}}){{\bf{R}}_{{{\cal E}_{ij}}}}}\\
      {{{\bf{t}}_{ij}} = \frac{1}{{\rm H}}\sum\limits_{k = 1}^K {\mathbb{I}{_{ij}}} \left( {{{\bf{p}}_k} - {{\bf{R}}_{ij}}{{\bf{q}}_k}} \right)}
      \end{array}} \right.
  \end{aligned}
  \label{[R|t]}
\end{equation}

Where $\rm H$ is the sum of the elements in the indicator function${{\mathbb{I}}_{ij}}\left( \cdot  \right)$.

\subsubsection{Obtain the global maximum consensus set}\label{sec:global maximum consensus}
In the correspondence graphs $({{\mathcal{G}}^{\mathcal{P}}},{{\mathcal{G}}^{\mathcal{Q}}})$,  the reliability ${{\mathcal{D}}_{\mathcal{G}}}(\mathcal{E}_{ij}^{\mathcal{P}\mathcal{Q}})\left| _{{{\mathcal{F}}_{1}}} \right.$ and ${{\mathcal{D}}_{\mathcal{G}}}(\mathcal{E}_{ij}^{\mathcal{P}\mathcal{Q}})\left| _{{{\mathcal{F}}_{2}}} \right.$ of each correspondence edge $(\mathcal{E}_{ij}^{\mathcal{P}},\mathcal{E}_{ij}^{Q})$ with their attached nodes $(V_{k}^{\mathcal{P}},V_{k}^{\mathcal{Q}})$ can be calculated according to the method in Section \ref{sec:Reliability of correspondece}. By comparing the reliability of each correspondence edge, we select the consensus set ${{\mathcal{I}}_{ij}}$ corresponding to the optimal matching edge as the global maximum consensus set $\mathcal{I}$. The alignment parameters based on the correspondence edge $(\mathcal{E}_{ij}^{\mathcal{P}},\mathcal{E}_{ij}^{Q})$ can be obtained with the process of calculating the reliability ${{\mathcal{D}}_{\mathcal{G}}}(\mathcal{E}_{ij}^{\mathcal{P}\mathcal{Q}})\left| _{{{\mathcal{F}}_{2}}} \right.$by selecting the constraint function ${{\mathcal{F}}_{2}}(\mathcal{E}_{ij}^{\mathcal{P}\mathcal{Q}},\mathcal{V}_{k}^{\mathcal{P}\mathcal{Q}})$, but the time complexity is $\mathcal{O}\left( K\log K \right)$. The time complexity of using the function ${{\mathcal{F}}_{1}}(\mathcal{E}_{ij}^{\mathcal{P}\mathcal{Q}},\mathcal{V}_{k}^{\mathcal{P}\mathcal{Q}})$ to calculate ${{\mathcal{D}}_{\mathcal{G}}}(\mathcal{E}_{ij}^{\mathcal{P}\mathcal{Q}})\left| _{{{\mathcal{F}}_{1}}} \right.$ is $\mathcal{O}\left( K \right)$, but it can’t finish the registration. In calculating the reliability of edges we follow the order of ${{\mathcal{D}}_{\mathcal{G}}}(\mathcal{E}_{ij}^{\mathcal{P}\mathcal{Q}})\left| _{{{\mathcal{F}}_{1}}} \right.$to ${{\mathcal{D}}_{\mathcal{G}}}(\mathcal{E}_{ij}^{\mathcal{P}\mathcal{Q}})\left| _{{{\mathcal{F}}_{2}}} \right.$. Since ${{\mathcal{F}}_{1}}$ is a more loose constraint function compared to ${{\mathcal{F}}_{2}}$, the two reliabilities of one correspondence edge have the following relationship: ${{\mathcal{D}}_{\mathcal{G}}}(\mathcal{E}_{ij}^{\mathcal{P}\mathcal{Q}})\left| _{{{\mathcal{F}}_{1}}} \right.>{{\mathcal{D}}_{\mathcal{G}}}(\mathcal{E}_{ij}^{\mathcal{P}\mathcal{Q}})\left| _{{{\mathcal{F}}_{2}}} \right.$. To obtain the global maximum consensus set, we need to compare the reliability of each edge. Given two edges $\mathcal{E}_{1}^{\mathcal{P}\mathcal{Q}}$ and $\mathcal{E}_{2}^{\mathcal{P}\mathcal{Q}}$,if the ${{\mathcal{D}}_{\mathcal{G}}}(\mathcal{E}_{1}^{\mathcal{P}\mathcal{Q}})\left| _{{{\mathcal{F}}_{2}}} \right.$ is greater than ${{\mathcal{D}}_{\mathcal{G}}}(\mathcal{E}_{2}^{\mathcal{P}\mathcal{Q}})\left| _{{{\mathcal{F}}_{1}}} \right.$, we skip calculate the ${{\mathcal{D}}_{\mathcal{G}}}(\mathcal{E}_{2}^{\mathcal{P}\mathcal{Q}})\left| _{{{\mathcal{F}}_{2}}} \right.$, which will reduce the calculation of ${{\mathcal{D}}_{\mathcal{G}}}(\mathcal{E}_{ij}^{\mathcal{P}\mathcal{Q}})\left| _{{{\mathcal{F}}_{2}}} \right.$ with higher time complexity and improve the efficiency. Based on this principle, a simple and efficient algorithm is designed for fast comparing the reliability of edges and obtaining the global maximum consensus set. The detail is shown in Algorithm 1.

\SetKwRepeat{Do}{do}{while}%
\begin{algorithm}
  \SetKwData{Left}{left}\SetKwData{This}{this}\SetKwData{Up}{up}
  \SetKwFunction{Union}{Union}\SetKwFunction{FindCompress}{FindCompress}
  \SetKwInOut{Input}{input}\SetKwInOut{Output}{output}

  \Input{Corresponding edge set:$\mathbf{E} = \{(\overrightarrow{\mathcal{E}}_{ij}^\mathcal{P},\overrightarrow{\mathcal{E}}_{ij}^Q)\} (i,j=1,2,...,K,i\ne j); \overrightarrow{\mathcal{E}}_{ij}^\mathcal{P}=\overrightarrow{{V_i}^\mathcal{P}{V_j}^\mathcal{P}},\overrightarrow{\mathcal{E}}_{ij}^\mathcal{Q}=\overrightarrow{{V_i}^\mathcal{Q}{V_j}^\mathcal{Q}};$ Node set for each edge pair:$\mathcal{V}_{\mathcal{E}_{ij}}^\mathcal{PQ} = \{({V_k}^\mathcal{P} ,{V_k}^\mathcal{Q})\}_{k=1}^K(k\ne i,j)\subseteq\mathcal{H'}$}

  \Output{Maximum consensus set: $\mathcal{I};$ transformation parameter:$\bm{R,t}$}

  \textbf{Initialize}: {Set $ O_{ij}=\varnothing, \mathcal{V}_{\mathcal{E}_{ij}}^\mathcal{'PQ} = \varnothing, \emph{bestcount} = 3$}\;

  \For{$each$ edge pair: ${\mathcal{E}}_{ij}^\mathcal{PQ}=({\mathcal{E}}_{ij}^\mathcal{P}, {\mathcal{E}}_{ij}^\mathcal{Q})$}
  {
    $(\bm{R}_{\overrightarrow{\mathcal{E}}_{ij}^\mathcal{PQ}}, \bm{t}_{\overrightarrow{\mathcal{E}}_{ij}^\mathcal{PQ}} ) \leftarrow alignEdgePair({\mathcal{E}}_{ij}^\mathcal{P},{\mathcal{E}}_{ij}^\mathcal{Q})$\;

    \For{$each$ node pair: ${V}_{k}^\mathcal{PQ}=({V}_{k}^\mathcal{P} , {V}_{k}^\mathcal{Q}) \in \mathcal{V}_{\mathcal{E}_{ij}}^\mathcal{PQ}$}
    {
        $({V}_{k}^\mathcal{'P} = {V}_{k}^\mathcal{P} , {V}_{k}^\mathcal{'Q} = \bm{R}_{\overrightarrow{\mathcal{E}}_{ij}^\mathcal{PQ}} \cdot{V_k}^\mathcal{Q} + \bm{t}_{\overrightarrow{\mathcal{E}}_{ij}^\mathcal{PQ}}) \rightarrow {V}_{k}^\mathcal{'PQ}=( {V}_{k}^\mathcal{'P},{V}_{k}^\mathcal{'Q}) $\;

        put ${V}_{k}^\mathcal{'PQ}$  into $\mathcal{V}_{\mathcal{E}_{ij}}^\mathcal{'PQ}$\;

        $m_{1}(k,1)\leftarrow  \mathcal{F}_{1}(\mathcal{E}_{ij}^{\mathcal{PQ}},\mathcal{V}_{ij}^\mathcal{'PQ}),m_1(k,1)\in \mathcal{M}_1$;\
    }

    $\mathcal{D_G}(\mathcal{E}_{ij}^\mathcal{PQ})|_{\mathcal{F}_1} = $ degree of $  \mathcal{M}_1$ \;

    \If{$\mathcal{D_G}(\mathcal{E}_{ij}^\mathcal{PQ})|_{\mathcal{F}_1} < \emph{bestcount}$}
    {
        \emph{continue} (return to 2)\;
    }

    \For{$each$ node pair: $ {V}_{k}^\mathcal{'PQ}=({V}_{k}^\mathcal{'P} , {V}_{k}^\mathcal{'Q}) \in \mathcal{V}_{\mathcal{E}_{ij}}^\mathcal{'PQ}$}
    {
        $\theta_k \in[\alpha_k,\beta_k] \leftarrow \mathcal{F}_2(\bm{\mathcal{E}}_{ij}^{\mathcal{PQ}},\mathcal{V}_{ij}^\mathcal{'PQ})$\;
    }

    $O_{ij} = \max\limits_\theta \sum\limits_{k=1}^K \mathbb{I}_{ij} (\theta_k \in [\alpha_k,\beta_k])$  (interval stabbing)\;

    $ m_2(k,1) \leftarrow  \mathbb{I}_{ij} \left(\cdot\right),m_2(k,1)\in \mathcal{M}_2$ \;

    $ \mathcal{D_G}(\mathcal{E}_{ij}^\mathcal{PQ})|_{\mathcal{F}_2} = $ degree of $  \mathcal{M}_2$ \;

    \If{$ \mathcal{D_G}(\mathcal{E}_{ij}^\mathcal{PQ})|_{\mathcal{F}_2} > \emph{bestcount}$}
    {
        $\emph{bestcount} = \mathcal{D_G}(\mathcal{E}_{ij}^\mathcal{PQ})|_{\mathcal{F}_2}+2 $\;

        $\mathcal{I} = \mathbb{I}_{ij} (\mathcal{V}_{\mathcal{E}_{ij}}^\mathcal{PQ}) \subseteq \mathcal{H'} $\;

        $\bm{R}=\bm{R}(\theta,\bm{\mathcal{E}_{ij}})\bm{R}_{\mathcal{E}_{ij}}, \bm{t} = \frac{1}{H} \sum\limits_{k=1}^K \mathbb{I}_{ij} (V_k^\mathcal{P} -\bm{R}V_k^\mathcal{Q}),H = \sum\limits_{k=1}^K \mathbb{I}_{ij}$\;
    }
  }
  \Return ${\mathcal{I},\bm{R},\bm{t}}$
  \caption{Maximum  consensus set based on edge reliability}
\end{algorithm}

The maximum consensus set $\tilde{\mathcal{I}}$ and the initial registration parameters $\mathbf{\tilde{R}},\mathbf{\tilde{t}}$ can be obtained simultaneously by executing Algorithm 1. In order to further optimize the registration parameters and obtain more accurate registration results, we transform the coordinates of the source key points in the original correspondence set $\mathcal{H} =\left\{ ({{\mathbf{p}}_{i}},{{\mathbf{q}}_{i}}) \right\}_{1}^{N}$:${{\mathbf{{q}'}}_{i}}=\mathbf{\tilde{R}}\cdot {{\mathbf{q}}_{i}}+\mathbf{\tilde{t}}$ and obtain the new correspondence $\mathcal{H}{ }'=\left\{ ({{\mathbf{p}}_{i}},{{{\mathbf{{q}'}}}_{i}}) \right\}_{1}^{N}$. If a correspondence $({{\mathbf{p}}_{i}},{{\mathbf{{q}'}}_{i}})\in \mathcal{H}{ }'$ satisfies the condition: $\left\| {{\mathbf{p}}_{i}}-{{{\mathbf{{q}'}}}_{i}} \right\|<\delta $, then $({{\mathbf{p}}_{i}},{{\mathbf{q}}_{i}})$ is considered as an inlier. Based on this principle, the original global maximal consensus set $\mathcal{I}\subseteq \mathcal{H} $ can be obtained by checking all correspondence in $\mathcal{H}$. Finally, according to $\mathcal{I}$, the rotation and translation matrixes $\mathbf{R},\mathbf{t}$ are reestimated using least-squares and SVD:
\begin{equation}
  \begin{aligned}
    \mathop {\operatorname{minimize}}\limits_{{\bf{R}},{\bf{t}}} \sum\limits_{i = 1}^N {\left\| {{{\bf{p}}_i} - \left( {{\bf{R}}{{\bf{q}}_i} + {\bf{t}}} \right)} \right\|} ,\left( {{{\bf{p}}_i},{{\bf{q}}_i}} \right) \in {\cal I} \subseteq {\cal H} 
  \end{aligned}
  \label{least-squares SVD}
\end{equation}

\section{EXPERIMENTS AND EVALUATIONS}\label{sec:experiments}
In the experiments, the proposed algorithm is implemented using C++ code and both simulations and challenging real-world data are adopted to evaluate the performance of our method.
\subsection{ Simulations}\label{sec:simulations}
\begin{figure*}[ht]
	\centering
	\includegraphics[width=2.0\columnwidth]{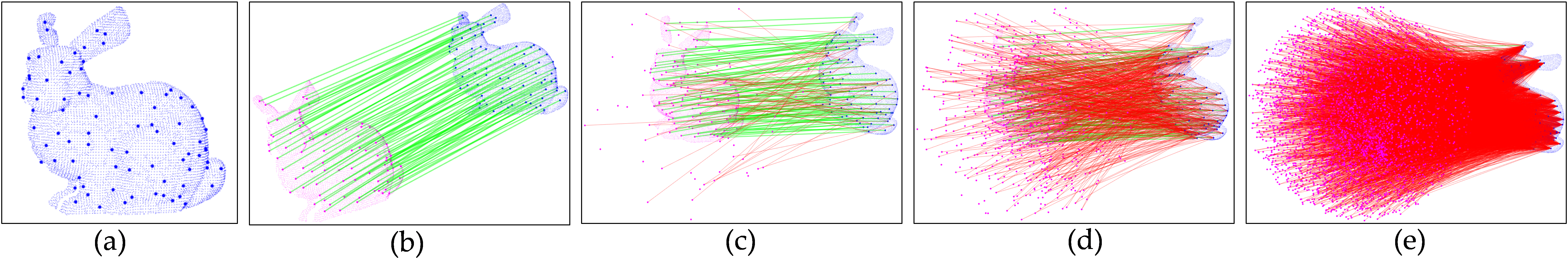}
	\caption{The generation of simulation data. (a) Key points in downsampling data, (b) inliers correspondence, (b) 50\% outliers are add, (c) 90\% outliers are add, and(d) 99\% outliers are add.}
  \label{Fig6}
\end{figure*}

\textbf{Testing Setup.} In the simulation experiment, we mainly test the performance of our algorithm in outlier removal. As shown in Fig. \ref{Fig6}(a), firstly, we downsampled the Stanford Bunny point cloud data by setting the resolution $\rho =0.002$. Then, we extracted ${{N}_{in}}$ = 80 key points using the ISS method as the source ${{\mathcal{Q}}_{in}}=\left\{ {{\mathbf{q}}_{i}} \right\}_{1}^{{{N}_{in}}}$. The correspondence target is ${{\mathcal{P}}_{in}}=\left\{ {{\mathbf{p}}_{i}} \right\}_{1}^{{{N}_{in}}}$, which is obtained by transformation operation: ${{\mathbf{p}}_{i}}=\left( \mathbf{R}{{\mathbf{q}}_{i}}+\mathbf{t} \right)$, where $\mathbf{R}$,$\mathbf{t}$ are the given transformation parameters, as shown in Fig. \ref{Fig6}(b). To make the simulation realistic, we add Gaussian noise $\mathcal{N}\left( 0,{{\rho }^{2}} \right)$ to the point set ${{\mathcal{P}}_{in}}$, obtaining an inlier set ${{\mathcal{H}}_{in}}=\left( {{\mathcal{P}}_{in}},{{\mathcal{Q}}_{in}} \right)$. To get outliers, we generate ${{N}_{out}}$ random points with the center of the point set ${{\mathcal{P}}_{in}}$ as the circle center and the diagonal of the bounding box of ${{\mathcal{P}}_{in}}$ as the radius, and randomly select ${{N}_{out}}$ point pairs in ${{\mathcal{Q}}_{in}}$ and ${{\mathcal{P}}_{out}}$ to form the outlier set ${{\mathcal{H}}_{out}}=\left( {{\mathcal{P}}_{out}},{{\mathcal{Q}}_{out}} \right)$. Finally, we merge ${{\mathcal{H}}_{in}}$ and ${{\mathcal{H}}_{out}}$ into the correspondence set $\mathcal{H}$, as shown in Fig. \ref{Fig6}(c)(d)(e).
To evaluate the effectiveness and robustness of the algorithm, we add ${N_{out}}$ = (50\%, 60\%, 70\%, 80\%, 90\%, 95\%, 96\%, 97\%, 98\%, 99\%) for testing, and each test is repeated 100 times. Some baseline as well as state-of-the-art outlier removal methods are selected for comparison, they are RANSAC \cite{[16]}, FGR \cite{[35]}, Gore \cite{[37]}, Teaser++ \cite{[40]}, and clipper \cite{[41]} respectively. See TABLE \ref{table1} for the setting of algorithm parameters. 

\begin{table*}[!t]
  \renewcommand{\arraystretch}{1.3}
  \caption{DETAILED SETTINGS OF THE COMPARED ALGORITHMS IN THE SIMULATION EXPERIMENT}
  \label{table1}
    \centering
    \begin{tabular}{l p{6.5cm} p{6.5cm}}
      \hline
      Method & Parameters & Implementations\\
      \hline
      RANSAC & Subset size: 3; confidence: 0.99; maximum number of iterations: $2\times {{10}^{4}}$; inlier threshold: $2\rho $. & \makecell[l]{C++ code; Single thread;\\https://github.com/PointCloudLibrary}\\

      FGR & Annealing rate: 1.4; maximum correspondence distance: $2\rho $; maximum number of iterations: 100. & C++ code; single thread;
      https://github.com/intel-isl/FastGlobalRegistration\\

      Gore & Lower bound: 0; repeat?: false; $\xi =\rho $. & \makecell[l]{C++ code; single thread;
      \\https://cs.adelaide.edu.au/aparra/project/gore}\\

      Teaser++ & Noise bound:$3\rho $; rotation max iterations:100; rotation gnc factor:1.4; rotation cost threshold:0.005. & \makecell[l]{C++ code; single thread;
      \\https://github.com/MIT-SPARK/TEASER-plusplus}\\

      FMP+BNB & $\epsilon =\rho $. & \makecell[l]{C++ code; single thread;
      \\https://github.com/ZhipengCai}\\

      Our method & K= 800; $\delta =\rho $. & \makecell[l]{C++ code; single thread;
      \\https://github.com/WPC-WHU/GROR}\\            
      \hline
    \end{tabular}
  \end{table*}

First, we evaluate the registration results, adopting widely used quality evaluation metrics: rotation error ${{\delta }_{\mathbf{R}}}$ and translation error ${{\delta }_{\mathbf{t}}}$ \cite{[10]}:
\begin{equation}
  \begin{aligned}
    \left\{ {\begin{array}{*{20}{l}}
        {\delta _{\bf{t}}} = {\left\| {{{\bf{t}}^t} - {{\bf{t}}^e}} \right\|_2}\\
        {\delta _{\bf{R}}} = \arccos \frac{{tr({{\bf{R}}^t}{{({{\bf{R}}^e})}^T}) - 1}}{2}
      \end{array}} \right.
  \end{aligned}
  \label{deltat&deltar}
\end{equation}

Where ${{\mathbf{R}}^{t}}$ and ${{\mathbf{t}}^{t}}$are rotation and translation matrix of reference (ground truth) parameter,  ${{\mathbf{R}}^{e}}$ and ${{\mathbf{t}}^{e}}$ are rotation and translation matrix of estimation parameter; $tr(\cdot )$ is the trace of a matrix. ${{\delta }_{\mathbf{R}}}$ measures the angular distance between ${{\mathbf{R}}^{t}}$ and ${{\mathbf{R}}^{e}}$, and ${{\delta }_{\mathbf{t}}}$ is the European distance between ${{\mathbf{t}}^{t}}$ and ${{\mathbf{t}}^{e}}$.
In addition, to evaluate the performance of outlier removal we adopt the metrics in clipper \cite{[37]} which calculate the Precision of outlier removal and the Recall rate of inliers. Let P be the number of ground truth inliers, TP is the number of ground truth inliers in the maximum consensus set $\mathcal{I}$ which is obtained by the algorithm, FP is the number of outliers in $\mathcal{I}$, and the metrics are calculated as follows:
\begin{equation}
  \begin{aligned}
    \left\{ {\begin{array}{*{20}{l}}
      {Precision = \frac{{TP}}{{TP + FP}}}\\
      {Recall = \frac{{TP}}{P}}
      \end{array}} \right.
  \end{aligned}
  \label{precision and recall}
\end{equation}

Finally, we calculate the average time of each test to evaluate the efficiency of the algorithm.
\begin{figure*}[!t]
	\centering
	\includegraphics[width=2.0\columnwidth]{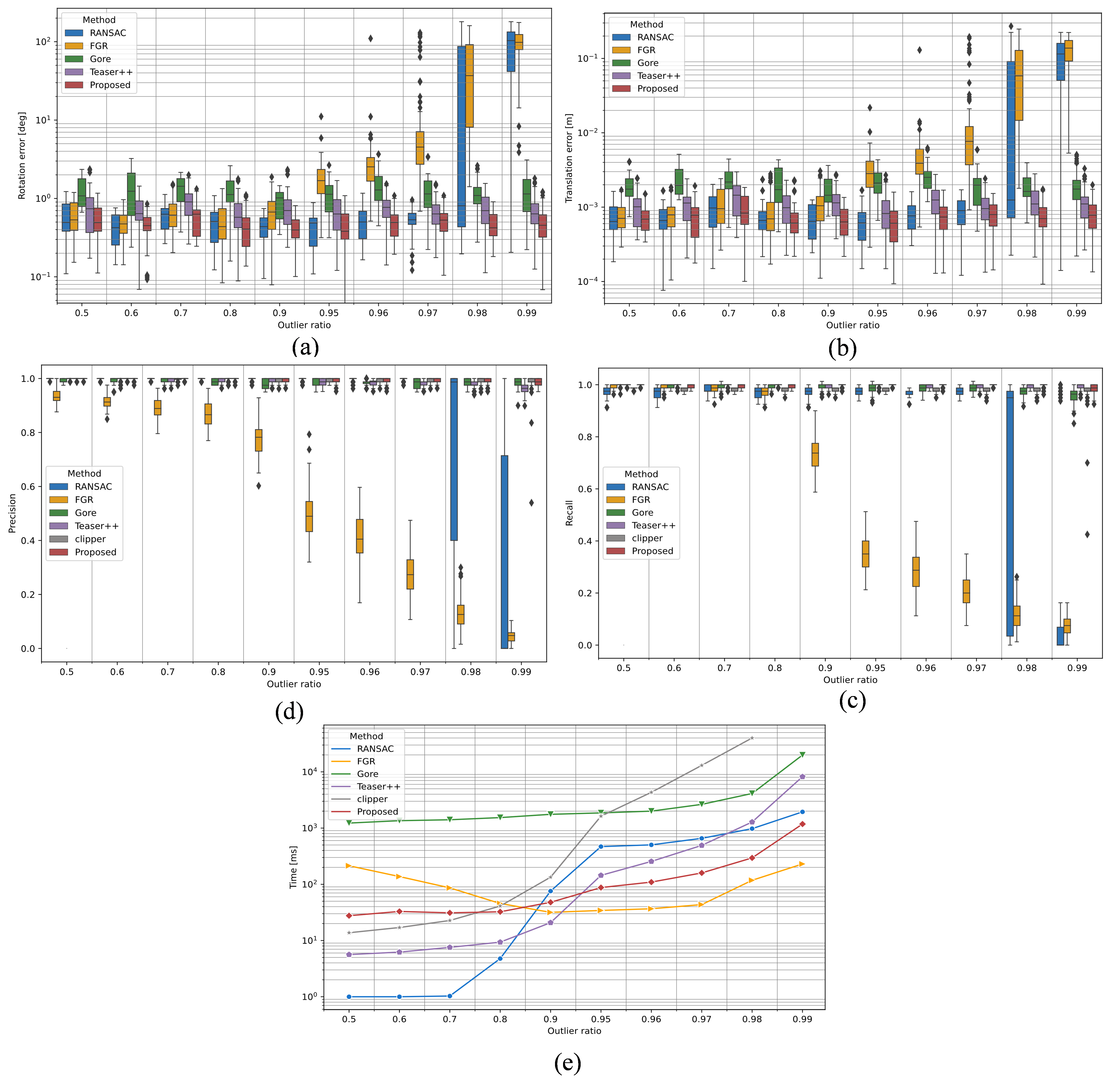}
	\caption{Performance of simulations experiment of algorithms. (a) Rotation performance, (b) translation performance, (b) Precision performance, (c) Recall performance, and(d) time performance.}
  \label{Fig7}
\end{figure*}

The metrics statistics are shown in Fig. \ref{Fig7}. It can be seen from the figure that: 1) The RANSAC \cite{[14]} algorithm performs well when the outlier ratio is less than 0.97, but when the outlier ratio is greater than 0.97, the metrics drop significantly. Because theoretically, they need more than 4 million trials to produce a good sample under 99\% outliers \cite{[34]}. 2) The FGR algorithm has poor performance in precision and recall. When the outlier ratio is less than 0.90, the FGR can obtain good rotation and translation performance, which is benefited from its optimization function. However, when the outlier ratio is greater than 0.90, although it still maintains high efficiency, the other four metrics become negative, which may be that the high outlier ratio makes the optimization function fail. 3) The Gore algorithm performs stably, when the outlier ratio is greater than 0.90, the rotation error of Gore is about 1.2°, the translation error is about 0.002. The precision and recall rate are all greater than 98\%. However, when the outlier ratio is low, the registration accuracy of the Gore is lower than that of RANSAC and FGR, and it takes 20s to complete the outlier removal and registration when the outlier ratio is greater than 0.99. 4) Teaser++ is an excellent point cloud registration method. It can be seen from Fig. \ref{Fig7} that Teaser++ maintains high registration accuracy and outlier removal performance in the whole experiment. However, when the outlier ratio is greater than 0.99, the whole process is about 8s on average, and the memory occupation is high when there are too many correspondences. 5) Clipper is a data association method with high outlier removal accuracy and inlier recall rate, but its efficiency is greatly reduced when there are more input correspondences.

As can be seen from Fig. \ref{Fig7}, when the outlier ratio exceeds 95\%, the performance of the proposed GROR is superior to the listed algorithms in rotation error, translation error, precision, and recall rate metrics. In terms of time performance, our algorithm is significantly better than other stable registration or outlier removal methods. The excellent performance of GROR in the simulation experiments could be illustrated by 1) The proposed method based on the node reliability with geometric constraint and the edge reliability theory with alignment constraints makes the average accuracy of outlier removal greater than 99\%, which is the best among all the algorithms.  2) The rotation matrix calculated directly by SVD based on exact inliers is more accurate, and thus obtain a more accurate translation vector, while Teaser++ uses a relaxed optimization method to calculate the rotation matrix, which affects the accuracy. 3) The presented optimal selection strategy based on node reliability removes most outliers which greatly reduces the number of correspondences. The edge reliability comparison method combining the loose and the tight constraint function could effectively reduce the high time-consuming operations. 

\subsection{Challenging real-world data}\label{sec:real-world}
In order to evaluate the performance of the proposed algorithm on real-world data, two datasets are selected for the experiment. One is the classical point cloud registration benchmark data: ETH dataset \cite{[47]}, which contains five scenes: Arch, Courtyard, Facade, Office, and Trees. Another set is WHU-TLS BENCHMARK \cite{[48]}, it contains 10 different scenes (subway station, high-speed railway platform, mountain, park, campus, residence, riverbank, heritage building, underground excavation, and tunnel) with varying point density, clutter, and occlusion. Completing the registration task of these 15 data sets is challenging. Since our algorithm is a pair-wise registration algorithm, the first and second scans of each scene from two data sets are chosen for the experiment to ensure generality. The details of each data set are shown in TABLE \ref{table2}.

In the challenging real-world experiment, we mainly test the performance of our algorithm in registration, so the rotation error ${{\delta }_{\mathbf{R}}}$, the translation error ${{\delta }_{\mathbf{t}}}$, and running time are selected to evaluate the performance. The RANSAC \cite{[14]}, a classical geometric constraint-based registration method K4PCS \cite{[49]}, FGR \cite{[50]}, Gore \cite{[33]}, Teaser++ \cite{[36]}, and a 4-DOF based method FMP+BNB \cite{[35]} are selected for comparison. The resolution of the downsampling for the ETH dataset is set to $\rho =0.1$m and the WHU-TLS benchmark is set to $\rho =0.2$ due to its large-scale data number. The detailed settings of the compared algorithms in the real-world experiment are shown in TABLE \ref{table3}.

\begin{figure*}[!t]
	\centering
	\includegraphics[width=2.0\columnwidth]{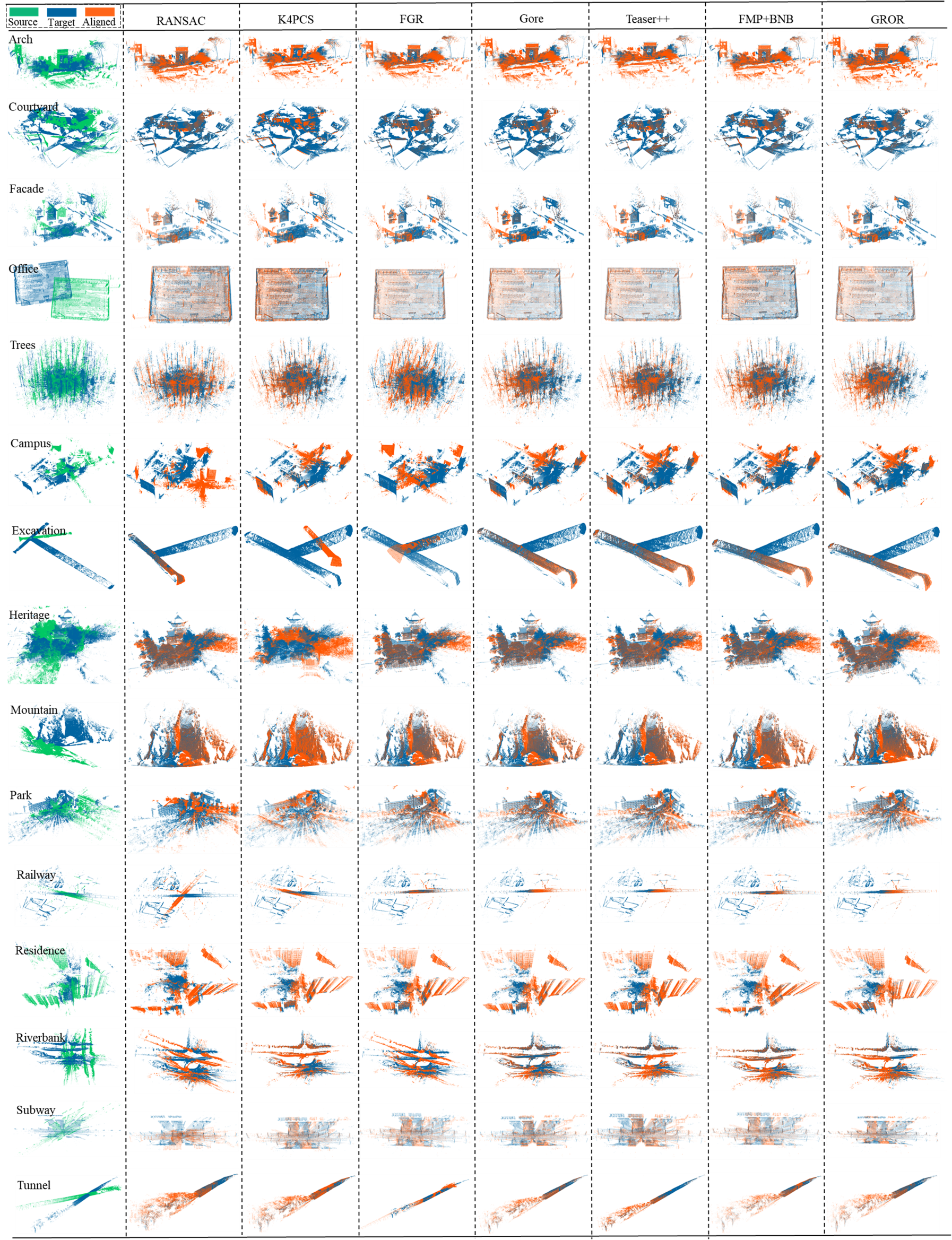}
	\caption{The visual performance of real-world data experiment of algorithms.}
  \label{Fig8}
\end{figure*}

Fig. \ref{Fig8} shows the registration results of all algorithms. Visually, the RANSAC and FMP+BNB algorithms have large translation errors in aligning the Railway data, and K4PCS performs poorly, with large rotation or translation errors on the Arch, Excavation, Heritage, Park, and Railway data. Gore and Teaser++ failed to complete the alignment of the Subway data. Our algorithm (GROR) completed the alignment for all 15 data sets effectively, which demonstrates the robustness of our algorithm.

\begin{figure*}[!t]
	\centering
	\includegraphics[width=2.0\columnwidth]{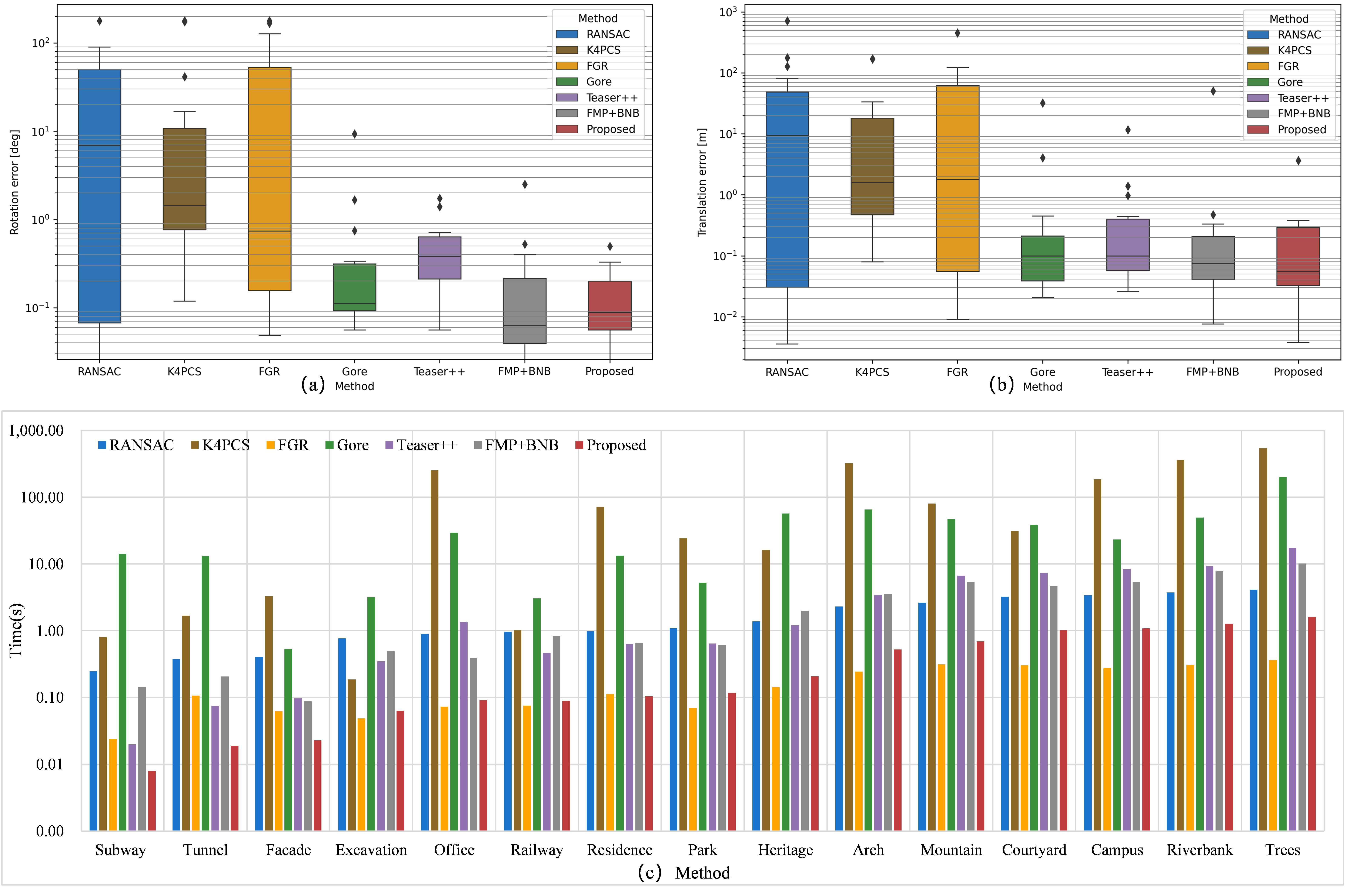}
	\caption{Quantitative evaluation results for 15 data sets (a) Box-plot of rotation error, (b) Box-plot of translation error, and(c)running time.}
  \label{Fig9}
\end{figure*}

To quantitatively evaluate the registration effect, we counted the rotation error ${{\delta }_{\mathbf{R}}}$: TABLE \ref{table4} and Fig. \ref{Fig9}(a), translation error ${{\delta }_{\mathbf{t}}}$ : TABLE \ref{table5} and Fig. \ref{Fig9}(b), and running time:  TABLE \ref{table6} and Fig. \ref{Fig9}(c). The RANSAC, K4PCS, and FGR algorithms are based on random sample consensus, and their registration results are unstable. From the quantitative evaluation results, we can see that 1) the registration performance of K4PCS is worse, the rotation error of 10 sets of data is more than 1 °, and the translation error of 9 sets of data is more than 1m. 2) The FGR algorithm is very efficient, but it is also very easy to fail when the outlier ratio is high or the registration scene is complex. There are 7 groups of data with rotation error of more than 1 °, and 8 groups of data with translation error of more than 1m. 3) The performance of the RANSAC is unstable, a total of 7 scene data were not correctly registered. Especially when the outlier ratio exceeds 99\%, the RANSAC is difficult to complete the registration task within the given iterations, and the time consuming to complete these iterations is at least an order of magnitude longer than our method GROR. 4) The registration result of the Gore algorithm is favorable, The alignment error of three data (Excavation, Subway, Tunnel) is more than 1°(${{\delta }_{\mathbf{R}}}$) or 1m (${{\delta }_{\mathbf{t}}}$), but the algorithm was not efficient overall, even when the number of correspondences is few or the outlier ratio is low. 5) The Teaser++ algorithm is capable of aligning most scenes, but the overall alignment accuracy is lower than that of RANSAC, Gore, FMP+BNB, and GROR, and the computation time increases significantly when the number of correspondences is large. 6) The FMP+BNB algorithm performs very well in real-world data, with low registration error overall, but it produces 2.510 ° rotation error and a large translation error of 50.61m in Railway data. Moreover, FMP+BNB has high computational efficiency, but the computation time increases when the number of correspondences is large. In addition, FMP+BNB is a 4-DOF registration algorithm and cannot accomplish 6-DOF alignment, which is why we did not show it in the simulation experiment. 

Finally, our algorithm GROR has a high translation and rotation registration accuracy, which is as good as FMP+BNB in the overall view, and our algorithm does not produce extreme errors, only a translation error of 3.636m appears in the Subway data. In the real-world data experiment, the biggest advantage of our algorithm is its efficiency. GROR can effectively complete the registration of all data and the efficiency of outlier removal for each data is higher than that of the state-of-the-art (Gore, Teaser++, and FMP+BNB), which is benefits from our outlier removal strategy based on the reliability of the correspondence graph.

\begin{table*}[]
  \renewcommand{\arraystretch}{1.3}
  \caption{MORE INFORMATION ABOUT EACH DATA SET}
  \label{table2}
  \centering
  \begin{tabular}{ccccccccc}
    \hline
  Dataset &
    Scene &
    \multicolumn{2}{l}{Scans} &
    Overlap   ratios &
    Points   ($10^6$) &
    \makecell{Key \\pointsnumber} &
    \makecell{Correspondence \\number}&
    Outlier   ratio \\
    \hline
  \multirow{5}{*}{\begin{tabular}[c]{@{}l@{}}ETH\\    \\ Dataset\end{tabular}} &
    \multicolumn{2}{l}{1-Arch} &
    1-2 &
    50\% &
    23.56-30.90 &
    8584-5208 &
    9827 &
    98.74\% \\
   & \multicolumn{2}{l}{2-Courtyard}  & 1-2   & 40\% & 13.32-18.80 & 4045-7546   & 13675 & 96.27\% \\
   & \multicolumn{2}{l}{3-Facade}     & 1-2   & 35\% & 25.08-15.25 & 686-1129    & 1614  & 96.84\% \\
   & \multicolumn{2}{l}{4-Office}     & 1-2   & 80\% & 10.72-10.71 & 1692-1681   & 3351  & 97.79\% \\
   & \multicolumn{2}{l}{5-Trees}      & 1-2   & 65\% & 19.63-19.60 & 12495-12719 & 17523 & 99.64\% \\
  \multirow{10}{*}{\begin{tabular}[c]{@{}l@{}}WHU-TLS   \\    \\ Dataset\end{tabular}} &
    \multicolumn{2}{l}{6-Campus} &
    1-2 &
    30\% &
    12.44-11.68 &
    8745-9991 &
    14364 &
    99.07\% \\
   & \multicolumn{2}{l}{7-Excavation} & 1-2   & 30\% & 40.01-39.65 & 893-3815    & 3123  & 97.37\% \\
   & \multicolumn{2}{l}{8-Heritage}   & 1-2   & 70\% & 31.48-28.50 & 4972-3619   & 5706  & 97.32\% \\
   & \multicolumn{2}{l}{9-Mountain}   & 1-2   & 85\% & 37.46-36.82 & 4804-4662   & 10838 & 93.84\% \\
   & \multicolumn{2}{l}{10-Park}      & 1-2   & 50\% & 5.66-5.07   & 2706-3096   & 4350  & 98.85\% \\
   & \multicolumn{2}{l}{11-Railway}   & 1-2   & 60\% & 6.16-5.06   & 1619-3061   & 3894  & 99.26\% \\
   & \multicolumn{2}{l}{12-Residence} & F2-F3 & 60\% & 5.56-7.36   & 3651-2294   & 4049  & 99.04\% \\
   & \multicolumn{2}{l}{13-Riverbank} & 1-2   & 80\% & 14.64-14.98 & 12719-12406 & 15564 & 99.70\% \\
   & \multicolumn{2}{l}{14-Subway}    & 1-3   & 65\% & 38.22-38.97 & 373-368     & 560   & 97.50\% \\
   & \multicolumn{2}{l}{15-Tunnel}    & 1-2   & 80\% & 22.36-22.42 & 910-722     & 1482  & 97.03\%\\
   \hline
  \end{tabular}
\end{table*}

\begin{table*}[]
  \renewcommand{\arraystretch}{1.3}
  \caption{DETAILED SETTINGS OF THE COMPARED ALGORITHMS IN THE REAL-WORLD EXPERIMENT}
  \label{table3}
  \centering
  \begin{tabular}{l p{6.5cm} p{6.5cm}}
    \hline
    Method & Parameters & Implementations\\
    \hline
    RANSAC &
    Subset size: 3; confidence: 0.99; maximum number of iterations: $2\times {{10}^{4}}$;inlier threshold: $2\rho $. &
   \makecell[l]{C++ code; Single thread; \\https://github.com/PointCloudLibrary}\\

   K4PCS &
    Delta: $\rho$; score threshold: 0.001. &
    \makecell[l]{C++ code; Multiple thread; \\ https://github.com/PointCloudLibrary}\\

    FGR &
    Annealing rate: 1.4; maximum correspondence distance: $2\rho $; maximum number of  iterations: 100. &
    \makecell[l]{C++ code; single thread; \\ https://github.com/intel-isl/FastGlobalRegistration}\\

    Gore &
    Lower bound: 0; repeat?: false; $\xi =\rho $. &
    \makecell[l]{C++ code; single thread;\\ https://cs.adelaide.edu.au/aparra/project/gore/}\\

   Teaser++ &
    Noise bound:$3\rho $; rotation max iterations:100; rotation gnc factor:1.4; rotation cost threshold:0.005. &
    \makecell[l]{C++ code; single thread;\\ https://github.com/MIT-SPARK/TEASER-plusplus} \\

  FMP+BNB &
    $\epsilon   =\rho $. &
    \makecell[l]{C++ code; single thread;\\ https://github.com/ZhipengCai/}\\

   Proposed &
    K= 800; $delta=rho$. &
    \makecell[l]{C++ code; single thread; \\ https://github.com/WPC-WHU/GROR}\\
    \hline

  \end{tabular}
\end{table*}

\begin{table*}[!t]
  \renewcommand{\arraystretch}{1.3}
  \caption{ROTATION ERROR ${\delta}_{\mathbf{R}}$(DEG)}
  \label{table4}
  \centering
  \begin{tabular}{cccccccccccccccc}
  \hline
  Method & 1     & 2     & 3     & 4     & 5     & 6     & 7     & 8     & 9     & 10    & 11    & 12    & 13    & 14    & 15    \\
  \hline
  RANSAC   & 0.079 & 0     & 0.088 & 179.1 & 13.78 & 90.00 & 0.056 & 0     & 0.048 & 42.37 & 58.13 & 6.915 & 16.70 & 90.00 & 0.692 \\
  K4PCS    & 4.700 & 0.631 & 2.782 & 0.776 & 2.189 & 1.444 & 41.52 & 174.6 & 0.119 & 179.0 & 16.90 & 0.759 & 0.220 & 1.260 & 1.313 \\
  FGR      & 0.097 & 0.048 & 0.291 & 0.407 & 127.0 & 167.9 & 86.78 & 0.048 & 0.056 & 0.217 & 0.739 & 1.110 & 19.72 & 8.238 & 179.3 \\
  Gore     & 0.137 & 0.056 & 0.338 & 0.193 & 0.088 & 0.084 & 0.748 & 0.097 & 0.088 & 0.290 & 0.112 & 0.108 & 0.112 & 9.305 & 1.669 \\
  Teaser++ & 0.427 & 0.063 & 0.236 & 1.402 & 0.384 & 0.188 & 0.281 & 0.108 & 0.056 & 0.619 & 0.714 & 0.500 & 0.253 & 1.735 & 0.659 \\
  FMP+BNB  & 0.040 & 0.040 & 0.101 & 0.400 & 0.125 & 0.137 & 0.063 & 0.523 & 0.056 & 0     & 2.510 & 0     & 0.048 & 0     & 0.293 \\
  Proposed & 0.056 & 0     & 0.175 & 0.325 & 0.125 & 0.148 & 0.079 & 0     & 0.048 & 0.329 & 0.055 & 0.227 & 0.084 & 0.496 & 0.101\\
  \hline
  \end{tabular}
\end{table*}

\begin{table*}[]
  \renewcommand{\arraystretch}{1.3}
  \caption{TRANSLATION ERROR ${\delta}_{\mathbf{t}}$(M)}
  \label{table5}
  \centering
  \begin{tabular}{cccccccccccccccc}
    \hline
  Method& 1     & 2     & 3     & 4     & 5     & 6     & 7     & 8     & 9     & 10    & 11    & 12    & 13    & 14    & 15    \\
  \hline
  RANSAC   & 0.011 & 0.025 & 0.030 & 12.04 & 10.77 & 177.4 & 0.137 & 0.030 & 0.004 & 9.506 & 127.3 & 14.93 & 82.65 & 719.3 & 0.043 \\
  K4PCS    & 1.711 & 1.004 & 1.595 & 0.104 & 0.689 & 2.000 & 169.8 & 33.63 & 0.162 & 28.47 & 172.1 & 0.721 & 0.253 & 7.729 & 0.079 \\
  FGR      & 0.026 & 0.021 & 0.086 & 0.079 & 9.655 & 123.0 & 453.7 & 0.055 & 0.009 & 0.057 & 70.84 & 1.798 & 99.61 & 53.93 & 36.27 \\
  Gore     & 0.047 & 0.031 & 0.135 & 0.024 & 0.025 & 0.271 & 4.032 & 0.021 & 0.087 & 0.100 & 0.046 & 0.154 & 0.448 & 32.16 & 0.106 \\
  Teaser++ & 0.118 & 0.026 & 0.065 & 0.091 & 0.039 & 0.364 & 1.399 & 0.059 & 0.038 & 0.100 & 0.206 & 0.439 & 0.971 & 11.74 & 0.056 \\
  FMP+BNB  & 0.017 & 0.035 & 0.008 & 0.090 & 0.048 & 0.334 & 0.211 & 0.062 & 0.019 & 0.073 & 50.61 & 0.074 & 0.203 & 0.471 & 0.111 \\
  Proposed & 0.026 & 0.026 & 0.055 & 0.051 & 0.038 & 0.297 & 0.285 & 0.039 & 0.005 & 0.084 & 0.068 & 0.310 & 0.382 & 3.636 & 0.051\\
  \hline
  \end{tabular}
\end{table*}

\begin{table*}[]
  \renewcommand{\arraystretch}{1.3}
  \caption{RUNNING TIME(S)}
  \label{table6}
  \centering
  \begin{tabular}{cccccccccccccccc}
  \hline
  Method  & 1     & 2     & 3     & 4     & 5     & 6     & 7     & 8     & 9     & 10    & 11    & 12    & 13    & 14    & 15    \\
  \hline
  RANSAC   & 2.310 & 3.235 & 0.406 & 0.895 & 4.113 & 3.414 & 0.771 & 1.388 & 2.630 & 1.095 & 0.965 & 0.986 & 3.760 & 0.172 & 0.378 \\
  K4PCS    & 324.8 & 31.27 & 3.318 & 254.4 & 541.2 & 185.7 & 0.187 & 16.32 & 80.30 & 24.52 & 1.035 & 71.50 & 362.9 & 0.812 & 1.688 \\
  FGR      & 0.246 & 0.305 & 0.062 & 0.073 & 0.363 & 0.278 & 0.049 & 0.144 & 0.314 & 0.070 & 0.076 & 0.113 & 0.308 & 0.024 & 0.107 \\
  Gore     & 65.53 & 38.67 & 0.532 & 29.38 & 201.6 & 23.24 & 3.186 & 57.21 & 47.00 & 2.250 & 3.050 & 13.36 & 49.59 & 14.18 & 13.11 \\
  Teaser++ & 3.423 & 7.394 & 0.098 & 1.358 & 17.41 & 8.394 & 0.349 & 1.215 & 6.720 & 0.646 & 0.469 & 0.636 & 9.344 & 0.020 & 0.075 \\
  FMP+BNB  & 3.567 & 4.632 & 0.088 & 0.391 & 10.19 & 5.403 & 0.495 & 1.944 & 5.397 & 0.611 & 0.828 & 0.654 & 7.906 & 0.145 & 0.207 \\
  Proposed & 0.526 & 1.023 & 0.023 & 0.092 & 1.617 & 1.088 & 0.063 & 0.180 & 0.694 & 0.118 & 0.089 & 0.105 & 1.281 & 0.008 & 0.019\\
  \hline
  \end{tabular}
\end{table*}

\begin{figure*}[t]
	\centering
	\includegraphics[width=2.0\columnwidth]{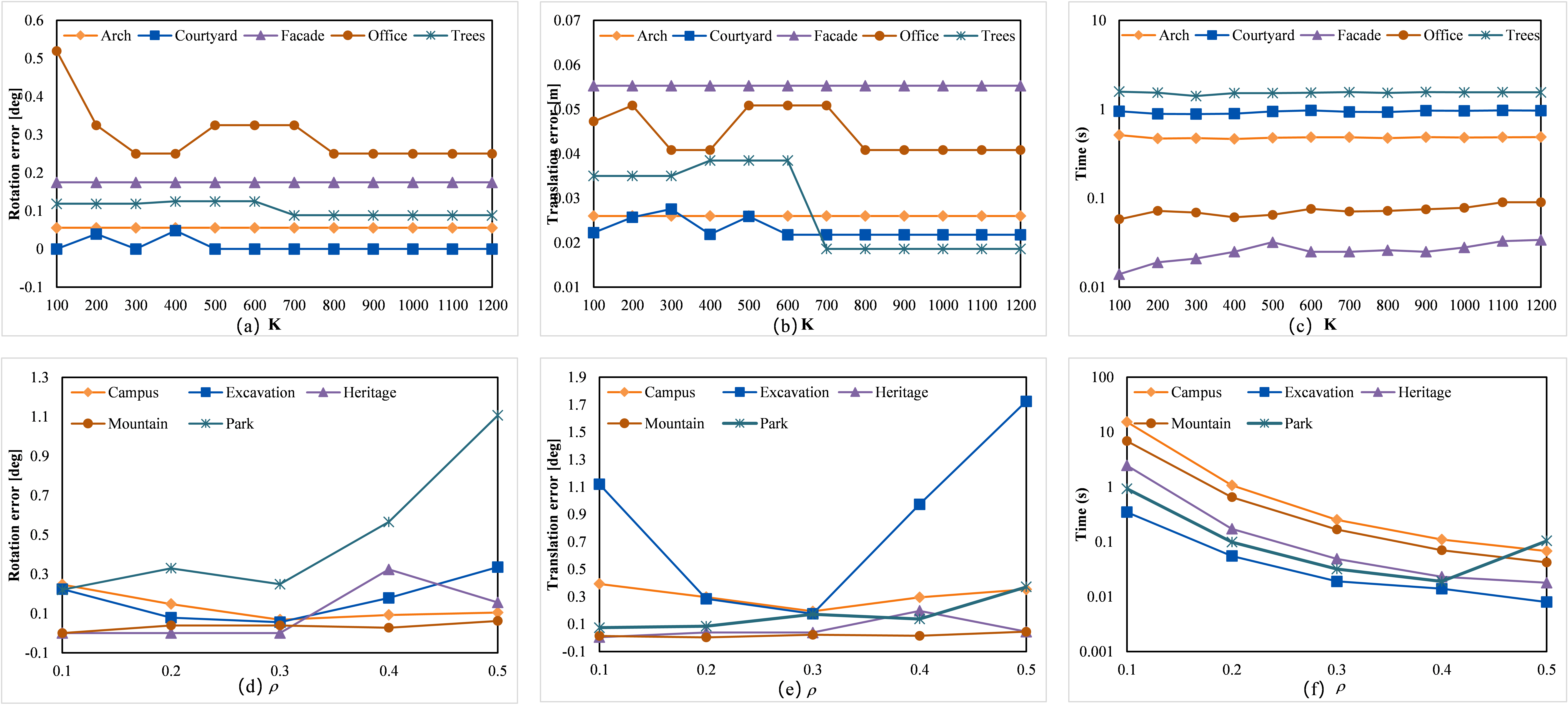}
	\caption{Sensitivity of parameters $K$ and $\rho $ (a) Rotation error for sensitivity test of $K$ , (b) translation error for sensitivity test of $K$, (c)running time for sensitivity test of $K$, (d) rotation error for sensitivity test of $\rho $, (e) translation error for sensitivity test of $\rho $, and (f) running time for sensitivity test of $\rho $.}
  \label{Fig10}
\end{figure*}

\subsection{Sensitivity of parameters}\label{sec:sensitivty of parameters}
GROR needs two parameters: the optimal selection parameter $K$ and the downsampling parameter $\rho $. The ETH dataset is used to test the sensitivity of the parameter $K$. We keep $\rho $ = 0.1, $K\in \left[ 100,1200 \right]$ and the interval is 100. The rotation and translation errors and running time evaluated by GROR at each $K$ is counted in Fig. \ref{Fig10}(a) - (c). It can be seen that all the rotation and translation errors are less than 1 ° and 0.1m. When $K\ge 800$, the rotation and translation errors tend to be stable. In terms of time performance, the running time increases slightly with the increase of $K$. So our algorithm is not sensitive to $K$ parameters. The parameter $K$ is set in the optimal selection strategy (OSS) which is used to select top $K$ reliable correspondence and improve the efficiency of the algorithm. An ablation experiment is conducted to evaluate the effect of the parameter $K$. Three test data-Facade, Heritage, and Trees-with different scales correspondence numbers (CN) are selected. The experimental results are shown in TABLE \ref{table7}:

\begin{table}[]
  \renewcommand{\arraystretch}{1.3}
  \caption{ABLATION EXPERIMENT RESULTS}
  \label{table7}
  \centering
  \begin{tabular}{cccc}
  \hline
  Data  & CN & \makecell{Running time \\without OSS(s)} & \makecell{Running time \\with OSS(s)}\\
  \hline
  Facade   & 1614  & 0.035 & 0023\\
  Heritage & 5706  & 0.549 & 0.180\\
  Trees    & 17523 & 5.493 & 1.617\\  
  \hline                 
  \end{tabular}
\end{table}

It can be seen from the table that OSS can effectively reduce the running time of the algorithm, and its effect is more obvious when there are more correspondences to be checked.  
Generally, setting the downsampling parameter $\rho $ to 0.1m can obtain a good registration result, but for larger-scale scene data such as WHU, it will produce a large number of correspondences, which will affect the performance of the algorithm. Therefore, in Section \ref{sec:real-world}, we set $\rho $ to 0.2m to enable all methods to complete the registration task within a comparable time frame. To test the sensitivity of $\rho $, $K$ is kept at 800 and $\rho \in \left[ 0.1,0.5 \right]$ with an interval of 0.1m. The rotation and translation errors and running times are counted in Fig. \ref{Fig10}(d) - (f). It can be seen that the parameter $\rho $ has an impact on the registration accuracy. On the whole, the registration accuracy will decrease with the increase of parameter$\rho $, but at the same time, there is a significant increase in efficiency. Meanwhile, the parameter $\rho $ can be set according to the size of the scale of the scenario, the data with large-scale scenarios need a larger $\rho $, the small-scale scenario needs a smaller $\rho $. The parameter $\rho $ affects the results of feature extraction and correspondences matching. Generally, all registration algorithms based on correspondence removal will be affected by the matching results.

\section{Conclusion}\label{sec:conclision}
In this paper, a new outlier removal strategy based on the reliability of correspondence graph for fast point cloud registration is proposed. The algorithm defines the concept of matching reliability of nodes and edges, designs the optimal candidate selection strategy according to the node reliability, and designs an algorithm for obtaining the global maximum consensus set according to the edge reliability, which can effectively and robustly remove outliers from a given correspondence, even if the outlier ratio exceeds 99\%. The 6-DOF transformation parameters of the pair-wise point cloud are calculated according to the accurate correspondences. The proposed algorithm is tested on simulation and real-world data respectively and compared with the classical baseline and state-of-the-art. The simulation experiments show that when the outlier ratio exceeds 95\%, the proposed algorithm has obvious advantages in the robustness of outlier removal. The real-world data experimental results show that the proposed method can complete the registration of all 15 test data. Compared with the classical registration algorithm, it has better scene practicability and is obviously better than the state-of-the-art in efficiency performance.

\ifCLASSOPTIONcompsoc
  \section*{Acknowledgments}
\else
  \section*{Acknowledgment}
\fi

The authors are grateful to the Editor-in-Chief, Associate Editor, and reviewers for their insightful and construc-tive comments. This work was supported in part by The National Key Research and Development Program of China under grant no. 2020YFD1100200; The Science and Technology Major Project of Hubei Province under Grant:
2021AAA010; National Natural Science Foundation of China (Grant No. 42171416).

\ifCLASSOPTIONcaptionsoff
  \newpage
\fi

\bibliographystyle{IEEEtran}

\bibliography{IEEEabrv,mycite}

\begin{thebibliography}{10}
\providecommand{\url}[1]{#1}
\csname url@samestyle\endcsname
\providecommand{\newblock}{\relax}
\providecommand{\bibinfo}[2]{#2}
\providecommand{\BIBentrySTDinterwordspacing}{\spaceskip=0pt\relax}
\providecommand{\BIBentryALTinterwordstretchfactor}{4}
\providecommand{\BIBentryALTinterwordspacing}{\spaceskip=\fontdimen2\font plus
\BIBentryALTinterwordstretchfactor\fontdimen3\font minus
  \fontdimen4\font\relax}
\providecommand{\BIBforeignlanguage}[2]{{%
\expandafter\ifx\csname l@#1\endcsname\relax
\typeout{** WARNING: IEEEtran.bst: No hyphenation pattern has been}%
\typeout{** loaded for the language `#1'. Using the pattern for}%
\typeout{** the default language instead.}%
\else
\language=\csname l@#1\endcsname
\fi
#2}}
\providecommand{\BIBdecl}{\relax}
\BIBdecl

\bibitem{[1]}
C.~C. Lin, Y.~C. Tai, J.~J. Lee, and Y.~S. Chen, ``A novel point cloud
  registration using 2d image features,'' \emph{Eurasip Journal on Advances in
  Signal Processing}, vol. 2017, pp. 1--11, 2017.

\bibitem{[2]}
S.~Szabó, P.~Enyedi, M.~Horváth, Z.~Kovács, P.~Burai, T.~Csoknyai, and
  G.~Szabó, ``Automated registration of potential locations for solar energy
  production with light detection and ranging (lidar) and small format
  photogrammetry,'' \emph{Journal of Cleaner Production}, vol. 112, pp.
  3820--3829, 2016.

\bibitem{[3]}
W.~Yao and U.~Stilla, ``Comparison of two methods for vehicle extraction from
  airborne lidar data toward motion analysis,'' \emph{IEEE Geoscience and
  Remote Sensing Letters}, vol.~8, pp. 607--611, 2011.

\bibitem{[4]}
M.~Huang, P.~Wei, and X.~Liu, ``An efficient encoding voxel-based segmentation
  (evbs) algorithm based on fast adjacent voxel search for point cloud plane
  segmentation,'' \emph{Remote Sensing}, vol.~11, p. 2727, 12 2019.

\bibitem{[5]}
P.~Kim, J.~Chen, and Y.~K. Cho, ``Slam-driven robotic mapping and registration
  of 3d point clouds,'' \emph{Automation in Construction}, vol.~89, pp. 38--48,
  2018.

\bibitem{[6]}
K.~Dong, S.~Gao, S.~Xin, and Y.~Zhou, ``Probability driven approach for point
  cloud registration of indoor scene,'' \emph{Visual Computer}, pp. 1--13,
  2020.

\bibitem{[7]}
\BIBentryALTinterwordspacing
J.~Závoti and J.~Kalmár, ``A comparison of different solutions of the
  bursa–wolf model and of the 3d, 7-parameter datum transformation,''
  \emph{Acta Geodaetica et Geophysica}, vol.~51, pp. 245--256, 2016. [Online].
  Available: \url{https://doi.org/10.1007/s40328-015-0124-6}
\BIBentrySTDinterwordspacing

\bibitem{[8]}
J.~Guo, J.~Shi, X.~Kong, and Z.~Liu, \emph{Foundation of geodesy}.\hskip 1em
  plus 0.5em minus 0.4em\relax Wuhan University Press, 2021.

\bibitem{[9]}
\BIBentryALTinterwordspacing
P.~Besl and N.~D. McKay, ``A method for registration of 3-d shapes,''
  \emph{IEEE Transactions on Pattern Analysis and Machine Intelligence},
  vol.~14, pp. 239--256, 2 1992. [Online]. Available:
  \url{http://ieeexplore.ieee.org/document/121791/}
\BIBentrySTDinterwordspacing

\bibitem{[10]}
\BIBentryALTinterwordspacing
P.~Wei, L.~Yan, H.~Xie, and M.~Huang, ``Automatic coarse registration of point
  clouds using plane contour shape descriptor and topological graph voting,''
  \emph{Automation in Construction}, p. 104055, 2021. [Online]. Available:
  \url{https://www.sciencedirect.com/science/article/pii/S0926580521005069}
\BIBentrySTDinterwordspacing

\bibitem{[11]}
R.~B. Rusu, N.~Blodow, and M.~Beetz, ``Fast point feature histograms (fpfh) for
  3d registration,'' 2009, pp. 3212--3217.

\bibitem{[12]}
S.~Salti, F.~Tombari, and L.~D. Stefano, ``Shot: Unique signatures of
  histograms for surface and texture description,'' \emph{Computer Vision and
  Image Understanding}, vol. 125, pp. 251--264, 2014.

\bibitem{[13]}
Z.~Dong, B.~Yang, Y.~Liu, F.~Liang, B.~Li, and Y.~Zang, ``A novel binary shape
  context for 3d local surface description,'' \emph{ISPRS Journal of
  Photogrammetry and Remote Sensing}, vol. 130, pp. 431--452, 2017.

\bibitem{[14]}
M.~A. Fischler and R.~C. Bolles, ``Random sample consensus: A paradigm for
  model fitting with applications to image analysis and automated
  cartography,'' \emph{Communications of the ACM}, vol.~24, pp. 619--638, 1981.

\bibitem{[15]}
S.~Mori, C.~Y. Sum, K.~Yamamoto, and J.-B. Cheng, ``Ransac-based darces: A new
  aooroach,'' vol.~21, 1999, pp. 1229--1234.

\bibitem{[16]}
\BIBentryALTinterwordspacing
R.~Huang, Y.~Xu, W.~Yao, L.~Hoegner, and U.~Stilla, ``Robust global
  registration of point clouds by closed-form solution in the frequency
  domain,'' \emph{ISPRS Journal of Photogrammetry and Remote Sensing}, vol.
  171, pp. 310--329, 2021. [Online]. Available:
  \url{https://www.sciencedirect.com/science/article/pii/S092427162030321X}
\BIBentrySTDinterwordspacing

\bibitem{[17]}
D.~Aiger, N.~J. Mitra, and D.~Cohen-Or, ``4-points congruent sets for robust
  pairwise surface registration,'' \emph{ACM Transactions on Graphics},
  vol.~27, pp. 1--10, 2008.

\bibitem{[18]}
E.~Xu, Z.~Xu, and K.~Yang, ``Using 2-lines congruent sets for coarse
  registration of terrestrial point clouds in urban scenes,'' \emph{IEEE
  Transactions on Geoscience and Remote Sensing}, p.~1, 2021.

\bibitem{[19]}
Y.~Xu, R.~Boerner, W.~Yao, L.~Hoegner, and U.~Stilla, ``Automated coarse
  registration of point clouds in 3d urban scenes using voxel based plane
  constraint,'' \emph{ISPRS Annals of the Photogrammetry, Remote Sensing and
  Spatial Information Sciences}, vol.~4, pp. 185--191, 2017.

\bibitem{[20]}
T.~NAKAMURA and S.~WAKITA, ``Robust global scan matching method using
  congruence transformation invariant feature descriptors and a geometric
  constraint between keypoints,'' \emph{Transactions of the Society of
  Instrument and Control Engineers}, vol.~51, pp. 309--318, 2015.

\bibitem{[21]}
D.~G. Lowe, ``Object recognition from local scale-invariant features,'' vol.~2,
  1999, pp. 1150--1157 vol.2.

\bibitem{[22]}
\BIBentryALTinterwordspacing
I.~Sipiran and B.~Bustos, ``Harris 3d: a robust extension of the harris
  operator for interest point detection on 3d meshes,'' \emph{The Visual
  Computer}, vol.~27, pp. 963--976, 2011. [Online]. Available:
  \url{https://doi.org/10.1007/s00371-011-0610-y}
\BIBentrySTDinterwordspacing

\bibitem{[23]}
Y.~Zhong, ``Intrinsic shape signatures: A shape descriptor for 3d object
  recognition,'' 2009, pp. 689--696.

\bibitem{[24]}
H.~Chen and B.~Bhanu, ``3d free-form object recognition in range images using
  local surface patches,'' \emph{Pattern Recognition Letters}, vol.~28, 2007.

\bibitem{[25]}
\BIBentryALTinterwordspacing
P.~W. Theiler, J.~D. Wegner, and K.~Schindler, ``Markerless point cloud
  registration with keypoint-based 4-points congruent sets,'' \emph{ISPRS Ann.
  Photogramm. Remote Sens. Spatial Inf. Sci.}, vol. II-5/W2, pp. 283--288, 10
  2013. [Online]. Available:
  \url{https://www.isprs-ann-photogramm-remote-sens-spatial-inf-sci.net/II-5-W2/283/2013/
  https://www.isprs-ann-photogramm-remote-sens-spatial-inf-sci.net/II-5-W2/283/2013/isprsannals-II-5-W2-283-2013.pdf}
\BIBentrySTDinterwordspacing

\bibitem{[26]}
\BIBentryALTinterwordspacing
Y.~Guo, F.~Sohel, M.~Bennamoun, M.~Lu, and J.~Wan, ``Rotational projection
  statistics for 3d local surface description and object recognition,''
  \emph{International Journal of Computer Vision}, vol. 105, pp. 63--86, 2013.
  [Online]. Available: \url{https://doi.org/10.1007/s11263-013-0627-y}
\BIBentrySTDinterwordspacing

\bibitem{[27]}
\BIBentryALTinterwordspacing
Z.~Jiao, R.~Liu, P.~Yi, and D.~Zhou, \emph{A Point Cloud Registration Algorithm
  Based on 3D-SIFT}, Z.~Pan, A.~D. Cheok, W.~Müller, M.~Zhang, A.~E. Rhalibi,
  and K.~Kifayat, Eds.\hskip 1em plus 0.5em minus 0.4em\relax Springer Berlin
  Heidelberg, 2019. [Online]. Available:
  \url{https://doi.org/10.1007/978-3-662-59351-6_3}
\BIBentrySTDinterwordspacing

\bibitem{[28]}
\BIBentryALTinterwordspacing
R.~Huang, W.~Yao, Z.~Ye, Y.~Xu, and U.~Stilla, ``Ridf: A robust
  rotation-invariant descriptor for 3d point cloud registration in the
  frequency domain,'' \emph{ISPRS Annals of the Photogrammetry, Remote Sensing
  and Spatial Information Sciences}, vol. V-2-2020, pp. 235--242, 2020.
  [Online]. Available:
  \url{https://www.isprs-ann-photogramm-remote-sens-spatial-inf-sci.net/V-2-2020/235/2020/}
\BIBentrySTDinterwordspacing

\bibitem{[29]}
Z.~J. Yew and G.~H. Lee, ``Rpm-net: Robust point matching using learned
  features,'' 6 2020.

\bibitem{[30]}
A.~Zeng, S.~Song, M.~Nießner, M.~Fisher, J.~Xiao, and T.~Funkhouser,
  ``3dmatch: Learning local geometric descriptors from rgb-d reconstructions,''
  vol. 2017-January, 2017.

\bibitem{[31]}
Z.~J. Yew and G.~H. Lee, ``3dfeat-net: Weakly supervised local 3d features for
  point cloud registration,'' vol. 11219 LNCS, 2018.

\bibitem{[32]}
J.~Li, Q.~Hu, and M.~Ai, ``Point cloud registration based on one-point ransac
  and scale-annealing biweight estimation,'' \emph{IEEE Transactions on
  Geoscience and Remote Sensing}, pp. 1--14, 2021.

\bibitem{[35]}
\BIBentryALTinterwordspacing
Z.~Cai, T.-J. Chin, A.~P. Bustos, and K.~Schindler, ``Practical optimal
  registration of terrestrial lidar scan pairs,'' \emph{ISPRS Journal of
  Photogrammetry and Remote Sensing}, vol. 147, pp. 118--131, 2019. [Online].
  Available:
  \url{https://www.sciencedirect.com/science/article/pii/S0924271618303125}
\BIBentrySTDinterwordspacing

\bibitem{[36]}
H.~Yang, J.~Shi, and L.~Carlone, ``Teaser: Fast and certifiable point cloud
  registration,'' \emph{IEEE Transactions on Robotics}, vol.~37, 2021.

\bibitem{[33]}
A.~P. Bustos and T.~J. Chin, ``Guaranteed outlier removal for point cloud
  registration with correspondences,'' \emph{IEEE Transactions on Pattern
  Analysis and Machine Intelligence}, vol.~40, pp. 2868--2882, 12 2018.

\bibitem{[34]}
J.~Li, ``A practical o(n2) outlier removal method for point cloud
  registration,'' \emph{IEEE Transactions on Pattern Analysis and Machine
  Intelligence}, 2021.

\bibitem{[37]}
P.~C. Lusk, K.~Fathian, and J.~P. How, ``Clipper: A graph-theoretic framework
  for robust data association,'' 2021, pp. 13\,828--13\,834.

\bibitem{[38]}
\BIBentryALTinterwordspacing
P.~H.~S. Torr and A.~Zisserman, ``Mlesac: A new robust estimator with
  application to estimating image geometry,'' \emph{Computer Vision and Image
  Understanding}, vol.~78, pp. 138--156, 2000. [Online]. Available:
  \url{https://www.sciencedirect.com/science/article/pii/S1077314299908329}
\BIBentrySTDinterwordspacing

\bibitem{[39]}
R.~Raguram, O.~Chum, M.~Pollefeys, J.~Matas, and J.-M. Frahm, ``Usac: A
  universal framework for random sample consensus,'' \emph{IEEE Transactions on
  Pattern Analysis and Machine Intelligence}, vol.~35, pp. 2022--2038, 2013.

\bibitem{[40]}
D.~Baráth, J.~Noskova, M.~Ivashechkin, and J.~Matas, ``Magsac++, a fast,
  reliable and accurate robust estimator,'' 2020, pp. 1301--1309.

\bibitem{[41]}
P.~Zhou, X.~Guo, X.~Pei, and C.~Chen, ``T-loam: Truncated least squares
  lidar-only odometry and mapping in real time,'' \emph{IEEE Transactions on
  Geoscience and Remote Sensing}, vol.~60, pp. 1--13, 2022.

\bibitem{[42]}
T.-J. Chin and D.~Suter, ``The maximum consensus problem: Recent algorithmic
  advances,'' \emph{Synthesis Lectures on Computer Vision}, vol.~7, 2017.

\bibitem{[43]}
\BIBentryALTinterwordspacing
K.~H. Lee, H.~Woo, and T.~Suk, ``Data reduction methods for reverse
  engineering,'' \emph{The International Journal of Advanced Manufacturing
  Technology}, vol.~17, pp. 735--743, 2001. [Online]. Available:
  \url{https://doi.org/10.1007/s001700170119}
\BIBentrySTDinterwordspacing

\bibitem{[44]}
F.~D. Wang, N.~Xue, Y.~Zhang, G.~S. Xia, and M.~Pelillo, ``A functional
  representation for graph matching,'' \emph{IEEE Transactions on Pattern
  Analysis and Machine Intelligence}, vol.~42, pp. 2737--2754, 11 2020.

\bibitem{[45]}
\BIBentryALTinterwordspacing
J.~van den Berg (https://math.stackexchange.com/users/91768/jur-van-den berg),
  ``Calculate rotation matrix to align vector a to vector b in 3d?''
  uRL:https://math.stackexchange.com/q/476311 (version: 2016-09-01). [Online].
  Available: \url{https://math.stackexchange.com/q/476311}
\BIBentrySTDinterwordspacing

\bibitem{[46]}
\BIBentryALTinterwordspacing
M.~de~Berg, M.~van Kreveld, M.~Overmars, and O.~C. Schwarzkopf,
  \emph{Computational Geometry}, M.~de~Berg, M.~van Kreveld, M.~Overmars, and
  O.~C. Schwarzkopf, Eds.\hskip 1em plus 0.5em minus 0.4em\relax Springer
  Berlin Heidelberg, 2000. [Online]. Available:
  \url{https://doi.org/10.1007/978-3-662-04245-8_1}
\BIBentrySTDinterwordspacing

\bibitem{[47]}
\BIBentryALTinterwordspacing
P.~W. Theiler, J.~D. Wegner, and K.~Schindler, ``Globally consistent
  registration of terrestrial laser scans via graph optimization,'' \emph{ISPRS
  Journal of Photogrammetry and Remote Sensing}, vol. 109, pp. 126--138, 2015.
  [Online]. Available:
  \url{https://www.sciencedirect.com/science/article/pii/S0924271615001987}
\BIBentrySTDinterwordspacing

\bibitem{[48]}
Z.~Dong, F.~Liang, B.~Yang, Y.~Xu, Y.~Zang, J.~Li, Y.~Wang, W.~Dai, H.~Fan,
  J.~Hyyppäb, and U.~Stilla, ``Registration of large-scale terrestrial laser
  scanner point clouds: A review and benchmark,'' pp. 327--342, 2020.

\bibitem{[49]}
\BIBentryALTinterwordspacing
P.~W. Theiler, J.~D. Wegner, and K.~Schindler, ``Keypoint-based 4-points
  congruent sets – automated marker-less registration of laser scans,''
  \emph{ISPRS Journal of Photogrammetry and Remote Sensing}, vol.~96, pp.
  149--163, 2014. [Online]. Available:
  \url{https://www.sciencedirect.com/science/article/pii/S0924271614001701}
\BIBentrySTDinterwordspacing

\bibitem{[50]}
Q.~Y. Zhou, J.~Park, and V.~Koltun, ``Fast global registration,'' vol. 9906
  LNCS.\hskip 1em plus 0.5em minus 0.4em\relax Springer Verlag, 2016, pp.
  766--782.

\end{thebibliography}

\end{document}